\RequirePackage{amsthm}

\documentclass[pdflatex,sn-apa]{sn-jnl}

\usepackage{graphicx}%
\usepackage{multirow}%
\usepackage{amsmath,amssymb,amsfonts}%
\usepackage{amsthm}%
\usepackage{mathrsfs}%
\usepackage[title]{appendix}%
\usepackage{xcolor}%
\usepackage{textcomp}%
\usepackage{manyfoot}%
\usepackage{booktabs}%
\usepackage{algorithm}%
\usepackage{algorithmicx}%
\usepackage{algpseudocode}%
\usepackage{listings}%
\usepackage{comment}

\usepackage{anyfontsize}
\usepackage[utf8]{inputenc} 
\usepackage[T1]{fontenc}    
\usepackage{hyperref}       
\usepackage{url}            
\usepackage{booktabs}       
\usepackage{amsfonts}       
\usepackage{nicefrac}       
\usepackage{microtype}      
\usepackage{xcolor}         
\usepackage{url}
\usepackage{hyperref}
\usepackage{array}
\usepackage{tabularx}
\usepackage{multirow, multicol}
\usepackage{adjustbox}
\usepackage{amssymb}
\usepackage{amsmath,amsthm}
\usepackage{algorithm}
\usepackage{algpseudocode}
\usepackage{graphicx}
\usepackage{amsbsy}
\usepackage{mathtools}
\usepackage{physics}
\usepackage{subcaption}
\usepackage{wrapfig}
\usepackage{rotating}

\newcolumntype{P}[1]{>{\centering\arraybackslash}p{#1}}
\newcolumntype{M}[1]{>{\centering\arraybackslash}m{#1}}
\newcommand{\rev}[1]{{\color{black}#1}}
\newcommand{\revf}[1]{{\color{black}#1}}

\newcommand{\com}[1]{{}} 
\newcommand{\dtl}[1]{{}} 
\newcommand{\bs}{\boldsymbol}

\theoremstyle{thmstyleone}%
\newtheorem{theorem}{Theorem}
\theoremstyle{thmstyletwo}%
\newtheorem{example}{Example}%
\newtheorem{remark}{Remark}%

\theoremstyle{thmstylethree}%
\theoremstyle{plain}
\newtheorem{lem}{Lemma}

\theoremstyle{definition}
\newtheorem{defn}{Definition}

\DeclareMathOperator{\vertex}{vert}
\DeclareMathOperator{\Int}{int}

\newcommand{\ceil}[1]{\left\lceil #1 \right\rceil}

\begin{document}

\title[Article Title]{Vector Optimization with Gaussian Process Bandits}


\author[1]{\fnm{İlter Onat} \sur{Korkmaz}}\email{onat.korkmaz@bilkent.edu.tr}

\author[1]{\fnm{Yaşar Cahit} \sur{Yıldırım}}\email{cahit.yildirim@bilkent.edu.tr}

\author[1]{\fnm{Çağın} \sur{Ararat}}\email{cararat@bilkent.edu.tr}
\author*[1]{\fnm{Cem} \sur{Tekin}}\email{cemtekin@ee.bilkent.edu.tr}

\affil*[1]{ \orgname{Bilkent University}, \orgaddress{ \city{Ankara}, \country{Türkiye}}}

\abstract{\rev{We study black-box vector optimization with Gaussian process bandits, where there is an incomplete order relation on objective vectors described by a polyhedral convex cone. Existing black-box vector optimization approaches either suffer from high sample complexity or lack theoretical guarantees. We propose \textit{Vector Optimization with Gaussian Process} (VOGP), an adaptive elimination algorithm that identifies Pareto optimal solutions sample efficiently by exploiting the smoothness of the objective function. We establish theoretical guarantees, deriving information gain-based and kernel-specific sample complexity bounds. Finally, we conduct a thorough empirical evaluation of VOGP and compare it with the state-of-the-art multi-objective and vector optimization algorithms on several real-world and synthetic datasets, emphasizing VOGP's efficiency (\textit{e.g.}, $\sim18\times$ lower sample complexity on average). We also provide heuristic adaptations of VOGP for cases where the design space is continuous and where the Gaussian process model lacks access to the true kernel hyperparameters. This work opens a new frontier in sample-efficient multi-objective black-box optimization by incorporating preference structures while maintaining theoretical guarantees and practical efficiency.}}




\maketitle

\section{Introduction}\label{sec1}

In diverse fields such as engineering, economics, and computational biology, the problem of identifying the best set of designs across multiple black-box objectives is a recurrent challenge. Often, the evaluation of a particular design is both expensive and noisy, leading to the need for efficient and reliable optimization methods. \rev{This issue is exemplified in fields like aerospace engineering and drug discovery, where each evaluation can involve costly experiments or simulations \citep{Obayashi_Yamaguchi_Nakamura_1997, Mukaidaisi_Vu_Grantham_Tchagang_Li_2022}.} A common framework for dealing with such multi-objective optimization problems is to focus on identifying Pareto optimal designs, for which there are no other designs that are better in all objectives. However, this approach can be restrictive, as it only permits a certain set of trade-offs between objectives where a disadvantage on one objective cannot be compensated by advantages on other objectives, i.e., the domination relation requires superiority in all dimensions.

A more comprehensive solution to this issue lies in vector optimization, where the partial ordering relations induced by convex cones are used. The core idea is to define a ``cone'' in the objective space that represents preferred directions of improvement. A solution is considered better than another if the vector difference between their objectives falls within this preference cone. The use of ordering cones in this way gives users a framework that can model a wide range of trade-off preferences \citep{jahn2009vector}.

\begin{figure}[h]
    \centering
    \includegraphics[scale=0.58]{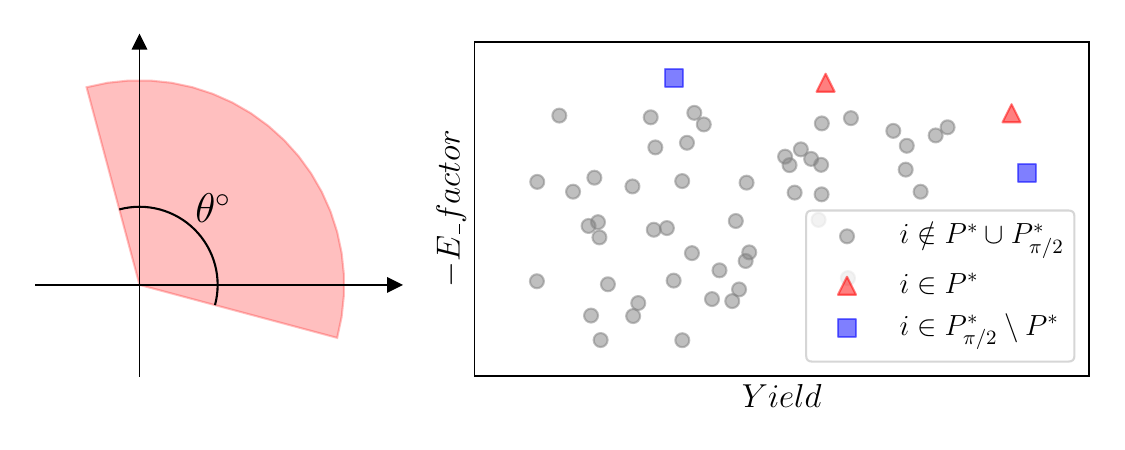}
    \caption{
    Vector optimization on nucleophilic aromatic substitution (SnAr) reaction data \citep{C6RE00109B}. (Left) The ordering cone. (Right) Cone-specific Pareto front ($P^*$) and its comparison to Pareto front of the componentwise order ($P^*_{\pi/2}$).
    }
    \label{fig:fourr}
    \vspace*{-0.3cm}
\end{figure}

Consider the challenge of optimizing the reactor parameters for nucleophilic aromatic substitution (SnAr) reaction, an important process in synthetic chemistry \citep{TAKAO2022294,GRANT20182437}. In this scenario, the goal is to maximize the space-time yield and minimize the waste-to-product ratio, i.e., the E-factor.\footnote{We provide experiments on this optimization task in Section~\ref{sec:exp}.} Consider two designs where one of the designs has slightly less space-time yield and a much lower E-factor. A reasonable preference would be to opt for the design that significantly reduces environmental impact, provided that the loss in space-time yield is not too substantial. Similarly, choosing designs that offer considerably higher space-time yield, albeit with a slightly increased environmental footprint, might also be justified. In vector optimization, this preference can be encoded with an obtuse cone (Figure~\ref{fig:fourr}). The regular Pareto optimality concept falls short in conveying such preferences as it requires superiority in all objectives for domination relations. In Figure~\ref{fig:fourr}, it can be seen that some of the designs (shown in red) in the Pareto front of traditional multi-objective order are not in the obtuse cone-based Pareto front, i.e., they are dominated by designs that are much better in one objective and slightly worse in the other.

\begin{figure}
\centering
 \includegraphics[scale=0.58]{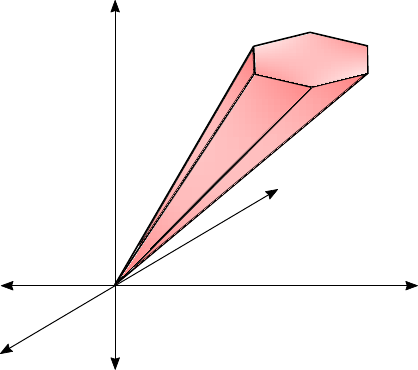}
 \caption{A 3D cone.}
\label{fig:six}
\end{figure}

The example we just discussed demonstrates vector optimization within a two-dimensional objective space, as this lower dimensionality allows for straightforward visualization. However, when extending to higher dimensions, the structure of the polyhedral cone may be more intricate with high number of faces. This is exemplified by Figure~\ref{fig:six}, which shows a polyhedral cone with six faces in three dimensions. Such cones are not uncommon in the literature. For example, \cite{hunt2010relative} provide a model of relative importance where allowable trade-offs between pairs of (possibly many) objectives are given. They provide a complete algebraic characterization of the cones corresponding to these kinds of preferences and show that they result in highly faceted polyhedral cones.

While there exists previous work addressing Pareto set identification with respect to an ordering cone in noisy environments with theoretical guarantees \citep{ararat2023vector,karagozlu2024learning}, they tend to require vast amounts of samples from the objective function, making them impractical in real-world scenarios. In this paper, we address the problem of identifying the Pareto set with respect to an ordering cone $C$ from a given set of designs with a minimum number of noisy black-box function queries. We propose {\em Vector Optimization with Gaussian Process} (VOGP), an ($\epsilon,\delta$)-{\em probably approximately correct} (PAC) adaptive elimination algorithm that performs black-box vector optimization by utilizing the modeling power of Gaussian processes (GPs). In particular, VOGP leverages confidence regions formed by GPs to perform cone-dependent, probabilistic elimination and Pareto identification steps, which adaptively explore the Pareto front with Bayesian optimization \citep{garnett_bayesoptbook_2023}. Similar to Pareto Active Learning (PAL) and $\epsilon$-PAL \citep{zuluaga2013active, JMLR:v17:15-047} for the multi-objective case (see next subsection), VOGP operates in a fully exploratory setting, where the least certain design is queried at each round. When VOGP is considered with the special case of componentwise order (positive orthant cone in vector optimization), it can be seen as a variant of PAL and $\epsilon$-PAL that can handle correlated objectives. We prove strong theoretical guarantees on the convergence of VOGP and derive information gain-based sample complexity bounds. Our main contributions can be listed as follows:
\begin{itemize}
    \item We propose VOGP, an ($\epsilon,\delta$)-PAC adaptive elimination algorithm that performs vector optimization by utilizing GPs.  This is the first work to provide a theoretical foundation for vector optimization within the framework of GP bandits.
    \item When the design set is finite, we theoretically prove that VOGP returns an ($\epsilon,\delta$)-PAC Pareto set with a cone-dependent sample complexity bound in terms of the maximum information gain. We also provide explicit kernel-specific sample complexity bounds.  
    \item We empirically show that VOGP satisfies the theoretical guarantees across different datasets and ordering cones. We also demonstrate that VOGP outperforms existing methods on vector optimization and its special case, multi-objective optimization. 
    \item We extend VOGP to work on continuous design sets by using tree-based adaptive discretization techniques. We empirically evaluate this extension in continuous design sets. \revf{We also study a heuristic variant of VOGP to handle the case of unknown GP kernel hyperparameters.} 
\end{itemize}

\section{Related Work}\label{sec:related}

There is considerable existing work on Pareto set identification that utilizes GPs \citep{10.1145/3195970.3196078,pmlr-v119-suzuki20a,10.1115/1.4046508,Mathern2021,Picheny2013MultiobjectiveOU,5949880, SVENSON2016250,candelieri2024fair}. Some of these works use information-theoretic acquisition functions. For instance, PESMO tries to sample designs that optimize the mutual information between the observation and the Pareto set, given the dataset \citep{pmlr-v48-hernandez-lobatoa16}. MESMO tries to sample designs that maximize the mutual information between observations and the Pareto front \citep{mesmo}. JES tries to sample designs that maximize the mutual information between observations and the joint distribution of the Pareto set and Pareto front \citep{jesmo}. Different from entropy search-based methods, some other methods utilize high-probability confidence regions formed by the GP posterior. For instance, PALS \citep{barracosa2022bayesian}, PAL and $\epsilon$-PAL are confidence region-based adaptive elimination algorithms that aim to categorize the input data points into three groups using the models they have learned: those that are Pareto optimal, those that are not Pareto optimal, and those whose statuses are uncertain \citep{zuluaga2013active, JMLR:v17:15-047}. In every iteration, these algorithms choose a potential design for evaluation with the aim of reducing the quantity of data points in the uncertain category. Thanks to GP-based modeling, a substantial number of designs can be explored by a single query. Assuming independent objectives, PAL and $\epsilon$-PAL work in the fixed confidence setting and have a variable sampling budget. On the other hand, entropy search-based methods are often investigated in fixed-budget setting without any theoretical guarantees on the accuracy of Pareto set identification. However, all the algorithms mentioned above try to identify the Pareto set with respect to the componentwise order. They are not able to incorporate user preferences encoded by ordering cones.

In the context of multi-objective Bayesian optimization, there is a limited amount of work that incorporates user preferences. \cite{NEURIPS2019_a7b7e4b2} propose MOBO-PC, a multi-objective Bayesian optimization algorithm which incorporates user preferences in the form of preference-order constraints. MOBO-PC uses expected Pareto hypervolume improvement weighted by the probability of obeying preference-order constraints as its acquisition function. \cite{7603186} propose a method that uses truncated expected hypervolume improvement and considers the predictive mean, variance, and the preferred region in the objective space. Some works employ preference learning \citep{10.1145/1102351.1102369,c0e55532870b4f48ba513ef249f152eb,10.1145/3449639.3459299,10.1145/3583133.3596342,ignatenko2021preference, bemporad2021global}, where the user interacts sequentially with the algorithm to learn the user preference \citep{Astudillo2019MultiattributeBO,Lin2022PreferenceEF}. \cite{Astudillo2019MultiattributeBO} employ Bayesian preference learning where the user's preferences are modeled as a utility function and they propose two novel acquisition functions that are robust to utility uncertainty. \cite{Lin2022PreferenceEF} consider various preference exploration methods while also representing user preferences with a utility function which is modeled with a GP. \cite{ahmadianshalchi2023preferenceaware} propose PAC-MOO, a constrained multi-objective Bayesian optimization algorithm which incorporates preferences in the form of weights that add up to one. PAC-MOO uses the information gained about the optimal constrained Pareto front weighted by preference weights as its acquisition function. \cite{khan2022efficient} propose a method that learns a utility function offline by using expert knowledge to avoid the repeated and expensive expert involvement.

\rev{In the broader context of gradient-free optimization, zeroth-order optimization techniques have been explored to handle optimization problems where gradient information is unavailable \citep{dang2024adaptive}. In the multi-objective case, numerous} studies employ evolutionary algorithms in order to estimate the Pareto front, thereby accomplishing this task through the iterative development of a population of evaluated designs \citep{10.1007/978-3-319-15892-1_3,996017,1583627,4358754,dang2023hybrid,dang2024generative}. Some of these methods use hypervolume calculation to guide their method \citep{7603187,BEUME20071653,6793178}. \rev{\cite{dang2024generative} incorporates generative adversarial networks into evolutionary algorithms to improve offspring diversity in the context of multimodal multi-objective optimization. \cite{dang2023hybrid} uses multi-objective evolutionary algorithms in federated learning for device selection, where multi-objective formulation helps balance resource efficiency and test accuracy.} There is a significant body of research that deals with the integration of user preferences into multi-objective evolutionary algorithms \citep{870272,ZHOU201132,branke2005integrating,7dd94870-9cf0-3ce8-815f-4a253b22d5f8,Li2008EvolutionaryMS,inprocefedings}. Using evolutionary algorithms, preferences can be articulated \textit{a priori}, \textit{a posteriori}, or interactively. Various methods for expressing preferences have been suggested, primarily encompassing techniques based on objective comparisons, solution ranking, and expectation-based approaches \citep{ZHOU201132}. There are also works that employ preference cones in multi-objective evolutionary algorithms. \cite{10.1007/978-3-642-19893-9_6} utilize polyhedral cones as a method of managing the resolution of the estimated Pareto front. \cite{FERREIRA2020115326} apply preference cone-based multi-objective evolutionary algorithm to optimize distributed energy resources in microgrids.

There are techniques that transform a multi-objective optimization problem into a single-objective problem by assigning weights to the objectives \citep{inproceedingss,1583627}. The transformed problem can be solved using standard single-objective optimization methods. ParEGO introduced in \cite{1583627} transforms the multi-objective optimization problem into a single-objective one by using Tchebycheff scalarization and solves it by Efficient Global Optimization (EGO) algorithm, which was designed for single-objective expensive optimization problems.

There are existing works that incorporate polyhedral structures to guide the design identification task. As an example, \cite{pmlr_v80_katz_samuels18a} introduce the concept of feasible arm identification. Their objective is to identify arms whose average rewards fall within a specified polyhedron, using evaluations that are subject to noise. \cite{ararat2023vector} propose a Na\"{i}ve Elimination (NE) algorithm for Pareto set identification with polyhedral ordering cones. They provide sample complexity bounds on the ($\epsilon$,$\delta$)-PAC Pareto set identification performance of this algorithm. However, their algorithm assumes independent arms and does not perform adaptive elimination. As a result, they only have worst-case sample complexity bounds for this algorithm. Experiments show that identification requires a large sampling budget which renders NE impractical in real-world problems of interest. 
\cite{karagozlu2024learning} introduce the PaVeBa algorithm, designed for identifying Pareto sets using polyhedral ordering cones. They also establish sample complexity bounds for the ($\epsilon$,$\delta$)-PAC Pareto set identification task within bandit settings. However, their analysis lacks theoretical guarantees when Gaussian processes are employed. 

\begin{table*}[h!]
\small
\caption{
\small
Comparison with Related Works
}
\centering
\label{tbl:compare_methods}
\scalebox{0.9}{
\begin{tabular}{M{0.33\linewidth}M{0.11\linewidth}M{0.17\linewidth}M{0.08\linewidth}M{0.06\linewidth}M{0.1\linewidth}}
\toprule
Paper & Cone-based preferences & Sample complexity bounds & ($\epsilon$,$\delta$)-PAC & Utilizes GPs & Continuous domain \\
\midrule
This work  & Yes & Yes & Yes & Yes & Yes \\
 \cite{karagozlu2024learning}  & Yes & Yes & Yes & No & No \\
 \cite{ararat2023vector}  & Yes & Yes & Yes & No & No \\
  \cite{jesmo}  & No & No & No & Yes & Yes \\
\cite{mesmo}  & No & No & No & Yes & Yes \\
\cite{pmlr-v48-hernandez-lobatoa16}  & No & No & No & Yes & Yes \\
 \cite{JMLR:v17:15-047}  & No & Yes & Yes & Yes & No \\

\bottomrule
\end{tabular}
}
\end{table*}

Table~\ref{tbl:compare_methods} compares our work with the most closely related prior works in terms of key \rev{aspects}. \rev{An important distinction is that our method supports cone-based preferences while also offering theoretical sample complexity guarantees in PAC framework. While \cite{karagozlu2024learning} and \cite{ararat2023vector} also incorporate cone-based preferences and provide sample complexity bounds, they do not utilize Gaussian processes and are restricted to discrete domains. In contrast, information theoretic methods such as PESMO \citep{pmlr-v48-hernandez-lobatoa16}, MESMO \citep{mesmo}, and JES \citep{jesmo} use Gaussian processes and operate in continuous domains but do not offer sample complexity guarantees, nor do they incorporate cone-based preferences. \cite{zuluaga2013active} employs Gaussian processes but does not incorporate cone-based preferences, and is also restricted to discrete domains.}

\section{Background and problem definition} \label{sec:PD}
\rev{The aim of this work is to incorporate polyhedral cones as preference structures to guide the multi-objective optimization, while maintaining theoretical guarantees and practical
efficiency. Formally,} we consider the problem of sequentially \revf{maximizing} a vector-valued function $\bs{f}: \mathcal{X} \rightarrow \mathbb{R}^M$ over a finite set of designs\footnote{Extension to continuous case is discussed in Section \ref{sec:exp}.} $\mathcal{X} \subset \mathbb{R}^D$ with respect to a polyhedral ordering cone $C \subset \mathbb{R}^M$. In each iteration $t$, a design point $x_t$ is selected, and a noisy observation is recorded as $\bs{y}_t = \bs{f}(x_t)+\bs{\nu}_t$. Here, the noise component $\bs{\nu}_t$ has the multivariate Gaussian distribution $\mathcal{N}\left(\bs{0},\sigma^2\bs{I}_{M\times M}\right)$, where $\sigma^2$ denotes the variance of the noise.  To define the optimality of designs, we use the partial order induced by $C$. \rev{A detailed table of notations can be found in Appendix \ref{secA1}.}

\subsection{Partial order induced by a cone} 

The partial order among designs is induced by a known polyhedral ordering cone $C\subseteq\mathbb{R}^M$, that is, a polyhedral closed convex cone whose interior, denoted by $\Int(C)$, is nonempty with $C\cap -C=\{\bs{0}\}$. Such a cone can be written as $C = \left\{\bs{x} \in \mathbb{R}^M \mid \bs{W} \bs{x} \geq \revf{\bs{0}}\right\}$, where $\bs{W}$ is a $N\times M$ matrix, $N$ is the number of halfspaces that define $C$, and $N\geq M$. Without loss of generality, the rows $\bs{w}_1^{\top}, \ldots, \bs{w}_N^{\top}$ are assumed to be unit normal vectors of these halfspaces with respect to the Euclidean norm $\|\cdot\|_2$. For $r\geq 0$, we write $\mathbb{B}(r)\coloneqq\{\bs{y}\in\mathbb{R}^M\mid \|\bs{y}\|_2\leq r\}$. We say that $\bs{y} \in \mathbb{R}^M$ weakly dominates ${{\bs{y}}^\prime}\in \mathbb{R}^M$ with respect to $C$ if their difference lies in $C$: ${{\bs{y}}^\prime} \preccurlyeq_C \bs{y} \Leftrightarrow \bs{y} -{{\bs{y}}^\prime} \in C$. By the structure of $C$, an equivalent expression is: 
\rev{
\begin{equation}\label{eq:conedomination}
{{\bs{y}}^\prime} \preccurlyeq_C \bs{y} \Leftrightarrow  \bs{w}_n^{\top}(\bs{y}-{{\bs{y}}^\prime}) \geq 0, \forall n\in [N]:=\{1,\ldots,N\} ~.
\end{equation}}

Moreover, we say that $\bs{y}$ strictly dominates ${{\bs{y}}^\prime}$ with respect to $C$ if their difference lies in $\Int(C)$: ${{\bs{y}}^\prime} \prec_C \bs{y} \Leftrightarrow \bs{y} -{{\bs{y}}^\prime} \in \Int(C)$. This is equivalent to: ${{\bs{y}}^\prime} \prec_C \bs{y} \Leftrightarrow  \bs{w}_n^{\top}(\bs{y}-{{\bs{y}}^\prime}) > 0, \forall n\in [N]$. Similarly, we write ${{\bs{y}}^\prime} \preccurlyeq_{C\setminus\{\bs{0}\}} \bs{y} \Leftrightarrow \bs{y} -{{\bs{y}}^\prime} \in C\setminus\{\bs{0}\}$ to exclude the case where the two vectors coincide. The Pareto set with respect to cone $C$ is the set of designs that are not weakly dominated by \revf{any other design} with respect to $C$ given as
\begin{align*}
    P^*:=P^*_C:=\{x \in \mathcal{X} \mid \nexists x^\prime \in \mathcal{X}: \bs{f}(x) \preccurlyeq_{C\setminus\{\bs{0}\}} \bs{f}(x^\prime)\}.
\end{align*}

\begin{figure}[H]
\centering
 \includegraphics[scale=0.68]{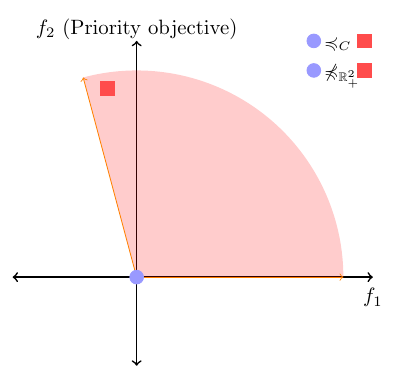}
 \caption{\rev{Visualization of a two-dimensional cone that prioritizes the $f_2$ objective. Unlike the componentwise order, the cone order establishes a domination relation by prioritizing $f_2$. The dominance relation under the preference cone is evident, as the difference of red square from blue circle lies inside the cone, confirming that the red square is preferred according to this ordering.}}
\label{fig:new}
\end{figure}

\rev{
\begin{example}
    Consider a scenario with two objectives $(M=2)$, denoted by $f_1$ and $f_2$. In this example, we assume that $f_2$ is prioritized over $f_1$, meaning the user is willing to trade off some amount of $f_1$ for a sufficient gain in $f_2$. Figure \ref{fig:new} illustrates the corresponding preference cone, where the red square is preferred over the blue circle. This preference structure differs from the componentwise order, under which these designs would be incomparable. 
\end{example}
}

\revf{
\begin{remark}
The relation $\preccurlyeq_C$ induced by the cone $C$ is a vector preorder on $\mathbb{R}^N$ that is non-total (incomplete). The choice $C=\mathbb{R}^M_+$ corresponds to the usual componentwise order in multi-objective optimization. In general, using a larger cones results in a smaller Pareto set; we refer the reader to \citet[Remark~2.3]{ararat2023vector} for possible uses of larger cones in small molecule drug discovery problem. In multi-asset markets with proportional transaction costs in finance, $C$ can be chosen as the so-called solvency cone determined by the transaction costs rates; see \cite{kabanov1999}. Cone-induced preorders form an important special class of preorders. The underlying scaling property, encoded by $C$ being a cone, is a strong assumption whose suitability should be justified in the particular application. For instance, in the case of solvency cones, the assumption that the transaction costs are incurred in proportion to the volume at a fixed rate justifies the use of a cone naturally.
\end{remark}
}


\revf{
\begin{remark}
    In this paper, we assume that the non-total order relation $\preccurlyeq_C$ is given a priori and the optimization task is performed with respect to it. A different strand of literature focuses on the problem of \emph{learning} the underlying order relation from data; see, e.g., 
    \citet{chu2005preference, chau2022learning, audiffren2017bandits, yue2012k}. Since the focus of the current paper is optimization with respect to a known vector preorder, the approach of these references is orthogonal to our work.
\end{remark}
}

\subsection{$(\epsilon,\delta)$-PAC Pareto set} 

We use the cone-dependent, direction-free suboptimality gaps defined in \cite{ararat2023vector}. The gap between designs $x,x^\prime\in\mathcal{X}$ is given as 
\begin{align*}
m(x, x^\prime) := \inf \{s \geq 0 \mid \exists \bs{u} \in \mathbb{B}(1) \cap C: \bs{f}(x)+s \bs{u} \notin \bs{f}(x^\prime)-\Int(C)\},
\end{align*}
the $-$ operation on sets is the Minkowski difference (vectors being treated as singletons). The suboptimality gap of design $x$ is defined as $\Delta_x^* := \max _{x^\prime \in P^*} m(x, x^\prime)$.

\begin{defn}\label{defn:success}
    Let $\epsilon > 0$, $\delta \in (0, 1)$. A random set $P \subseteq \mathcal{X}$ is called an \emph{$(\epsilon, \delta)$-PAC Pareto set} if the subsequent conditions are satisfied at least with probability $1-\delta$:\\
    (i) $\bigcup_{x \in P}\left(\bs{f}(x)+ \mathbb{B}(\epsilon) \cap C-C\right) \supseteq \bigcup_{x \in P^*}\left(\bs{f}(x)-C\right)$; \quad (ii) $\forall x \in P \setminus P^*\colon \Delta_x^* \leq 2\epsilon$.
\end{defn}

\begin{remark}
    Condition (i) of Definition \ref{defn:success} is equivalent to the following: For every $x^*\in P^*$, there exist $x\in P$ and $\bs{u} \in \mathbb{B}(\epsilon) \cap C$ such that $\bs{f}(x^*) \preccurlyeq_C \bs{f}(x)+\bs{u}$. Even though certain designs in $P$ may not be optimal, condition (ii) of Definition \ref{defn:success} limits the extent of the inadequacies in these designs. Consequently, it ensures the overall quality of all the designs produced.
\end{remark}

Our aim is to design an algorithm that returns an $(\epsilon,\delta)$-PAC Pareto set $\hat{\mathcal{P}} \subseteq \mathcal{X}$ with as little sampling from the expensive objective function $\bs{f}$ as possible.

\subsection{Modeling of the objective function}

 We model the objective function $\bs{f}$ as a realization of an $M$-output GP with zero mean and a positive definite covariance function $\bs{k}$ with bounded variance: $k^{jj}(x,x)\leq 1$ for every $x\in\mathcal{X}$ and $j\in[M]$. Let $\tilde{x}_i$ be the $i$th design observed and $\bs{y}_i$ the corresponding observation. The posterior distribution of $\bs{f}$ conditioned on the first $t$ observations is that of an $M$-output GP with mean ($\bs{\mu}_t$) and covariance ($\bs{k}_t$) functions given below, where $\bs{k}_{t}(x)=\left[\bs{k}\left(x, x_1\right), \ldots, \bs{k}\left(x, x_t\right)\right] \in \mathbb{R}^{M \times M t}$, $\bs{y}_{[t]}=\left[\bs{y}_1^{\top}, \ldots, \bs{y}_t^{\top}\right]^{\top}$, $\bs{K}_{t} =(\bs{k}\left(x_i, x_j\right))_{i,j \in [t]}$, and $\bs{I}_{Mt}$ is the $Mt\times Mt$ identity matrix:
\begin{align} \label{eq:posterior}
    &\bs{\mu}_t(x)=\bs{k}_{t}(x)\left(\bs{K}_{t}+\sigma^2\bs{I}_{Mt}\right)^{-1} \bs{y}_{[t]}^{\top}, \\
     & \bs{k}_t\left(x, x^{\prime}\right)=\bs{k}\left(x, x^{\prime}\right)-\bs{k}_{t}(x)\left(\bs{K}_{t}+\sigma^2\bs{I}_{Mt}\right)^{-1} \bs{k}_{t}\left(x^{\prime}\right)^{\top}~. 
\end{align}

Our sample complexity results will depend on the difficulty of learning $\bs{f}$, which is characterized by its maximum information gain. 

\begin{defn}
    The maximum information gain at round $t$ is defined as $\gamma_t\coloneqq \max_{A \subseteq \mathcal{X}:|A|=t} \mathrm{I}(\bs{y}_A ; \bs{f}_A)$, where $\bs{y}_A$ is the collection of observations corresponding to the designs in $A$, $\bs{f}_A$ is the collection of the corresponding function values, and $\mathrm{I}(\bs{y}_A;\bs{f}_A)$ is the mutual information between the two.
\end{defn}

\section{Method}

We propose {\em Vector Optimization with Gaussian Process} (VOGP), an ($\epsilon,\delta$)-PAC adaptive elimination algorithm that performs vector optimization by utilizing the modeling power of GPs (see Algortihm \ref{alg:highlevel}). VOGP takes as input the design set $\mathcal{X}$, cone $C$, accuracy parameter $\epsilon$, and confidence parameter $\delta$.
VOGP adaptively classifies designs as suboptimal, Pareto optimal, and uncertain. Initially, all designs are put in the set of uncertain designs, denoted by $\mathcal{S}_t$. Designs that are classified as suboptimal are discarded and never considered again in comparisons, making it sample-efficient. Designs that are classified as Pareto optimal are moved to the predicted Pareto set $\hat{\mathcal{P}}_t$. \rev{VOGP tracks the set of active designs by assigning them to $\mathcal{A}_t$ before discarding and $\mathcal{W}_t$ afterwards.} VOGP terminates and returns $\hat{\mathcal{P}}=\hat{\mathcal{P}}_t$ when the set of uncertain designs becomes empty. Decision-making in each round is divided into four phases named {\em modeling}, {\em discarding}, {\em Pareto identification}, {\em evaluating}. Below, we describe each phase in detail.

\begin{algorithm}[H]
\caption{VOGP}\label{alg:highlevel}
\begin{algorithmic}
\State \textbf{Input:} Design set $\mathcal{X}$, $\epsilon \geq 0$, $\delta \in (0,1)$, polyhedral ordering cone $C$, \revf{kernel $\bs{k}$}
\State \textbf{Initialize:} \rev{Predicted Pareto set} $\mathcal{P}_1 = \emptyset$,  \rev{set of uncertain designs} $ \mathcal{S}_1 = \mathcal{X}$, \rev{sets of active designs $\mathcal{A}_1 =\emptyset$ and $ \mathcal{W}_1 = \emptyset$,} $\boldsymbol{R}_0(x) = \mathbb{R}^M$ for each $x \in \mathcal{S}_1$, $t = 1$
\State \textbf{Compute:} Accuracy vector $\boldsymbol{u}^* \in C$ \rev{given in Definition \ref{defn:accuracy}}

\While{$\mathcal{S}_t \neq \emptyset$}
    \State Run \Call{Modeling}{$\mathcal{S}_t, \mathcal{P}_t$} in Algorithm \ref{alg:modeling}
    \State Run \Call{Discarding}{$\mathcal{S}_t,\mathcal{A}_t$} in Algorithm \ref{alg:discarding}
    \State Run \Call{ParetoIdentification}{$\mathcal{S}_t, \mathcal{P}_t$} in Algorithm \ref{alg:pareto}
    \State Run \Call{Evaluating}{$\mathcal{S}_t, \mathcal{W}_t$} in Algorithm \ref{alg:evaluating}
    \State $\mathcal{P}_{t+1} = \mathcal{P}_{t}$, $\mathcal{S}_{t+1} = \mathcal{S}_{t}$
    \State $t = t + 1$
\EndWhile

\State \textbf{return} Predicted Pareto set $\hat{\mathcal{P}} = \mathcal{P}_{t}$
\end{algorithmic}
\end{algorithm}

\revf{


\begin{remark}
Our theoretical results, similar to other theoretical works related to ours \citep{srinivas2012information, zuluaga2013active}, assume that the kernel is known and fixed. In practice, however, the kernel may be unknown and must be estimated from data concurrently with the decision-making process; see, e.g., \citet{berkenkamp2019no, ziomek2024bayesian}. We discuss this practical case in Section~\ref{app:unkown}, where we provide an empirical adaptation of Algorithm~\ref{alg:highlevel} for unknown hyperparameters.
\end{remark}
}

\subsection{Modeling} This phase is summarized in Algorithm \ref{alg:modeling}. VOGP uses the GP posterior means and variances of designs to define confidence regions in the form of $M$-dimensional hyperrectangles which are scaled by $\beta_t$, a function of the confidence parameter $\delta$. The probability that these hyperrectangles include the true objective values of the designs is at least $1-\delta$.
Using these confidence hyperrectangles allows for the formulation of probabilistic success criteria and the quantification of uncertainty with designs. At each round $t$, the hyperrectangle
\begin{align}\label{eq:hyperrec}
    \boldsymbol{Q}_t(x) = \{ &\boldsymbol{y} \in \mathbb{R}^M \mid \mu^j_t(x) - \beta_t^{1/2} \sigma^j_t(x) \leq y^j \leq \mu^j_t(x) + \beta_t^{1/2} \sigma^j_t(x), \forall j \in [M] \}
\end{align}
is computed for design $x$, where $\beta_t = 2 \ln (M\pi^2 |\mathcal{X}| t^2/(3\delta))$.

\begin{algorithm}[H]
\caption{Modeling Subroutine}\label{alg:modeling}
\begin{algorithmic}
\State \rev{\textbf{Compute:} Set of active designs $\mathcal{A}_t = \mathcal{S}_t \cup \mathcal{P}_t$};
\For{$x \in \mathcal{A}_t$}
    \State Obtain GP posterior for $x$: $\boldsymbol{\mu}_{t}(x)$ and $\boldsymbol{\sigma}_{t}(x)$ \rev{given in Equation \ref{eq:posterior}};
    \State Construct confidence hyperrectangle $\boldsymbol{Q}_{t}(x)$ \rev{given in Equation  \ref{eq:hyperrec}};
    \State $\boldsymbol{R}_{t}(x) = \boldsymbol{R}_{t-1}(x) \cap \boldsymbol{Q}_{t}(x)$;
\EndFor
\end{algorithmic}
\end{algorithm}

Moreover, all the calculated hyperrectangles up to round $t$ are intersected cumulatively to obtain the cumulative confidence hyperrectangle $\boldsymbol{R}_{t}(x)$ of design $x$.

\subsection{Discarding} 
\begin{figure*}[!t]
    \centering
    \includegraphics[scale=0.7]{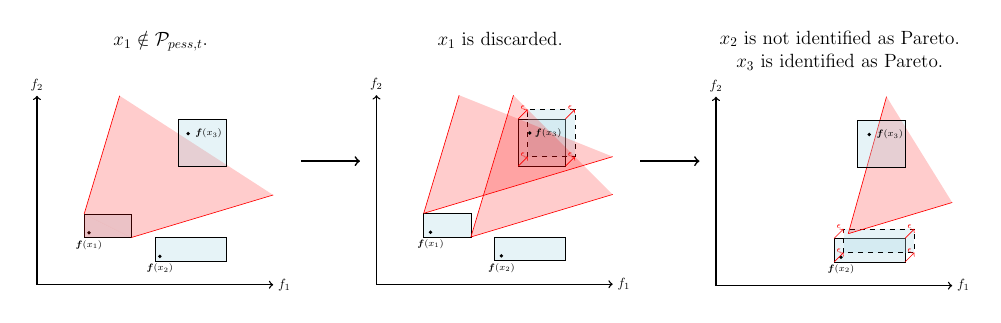}
    \vspace{-0.3cm}
    \caption{
    (Left, Middle) Visualization of discarding, and (Right) Pareto identification phases of VOGP in two dimensional objective space. Note that all boxes correspond to the $\boldsymbol{R}_t(x_i)$ for corresponding true objective values $\boldsymbol{f}(x_i)$ for $i=1,2,3$. In this simple case, in the middle plot, the intersection of the two cones are equivalent to the intersection of all cones coming out of $\boldsymbol{R}_t(x_1)$.
    \vspace{-0.25cm}
    }
    \label{fig:enter-label}
\end{figure*}

This phase is summarized in Algorithm \ref{alg:discarding}. VOGP discards undecided designs that are dominated with respect to $C$ with high probability. The designs that are to be discarded with low probability are identified and are used to form the pessimistic Pareto set $\mathcal{P}^{(C)}_{\text{pess},t}(\mathcal{A}_t)$ as defined below. The illustration of discarding phase can be seen in Figure~\ref{fig:enter-label}(Left, Middle).

\begin{defn}[Pessimistic Pareto Set]\label{defn:pess}
  Let $t \geq 1$ and let $D \subseteq \mathcal{A}_t$ be a nonempty set of nodes. The \emph{pessimistic Pareto set} of $D$ with respect to $C$ at round $t$, denoted by $\mathcal{P}^{(C)}_{\text{pess},t}(D)$, is the set of all nodes $x \in D$ for which there is no node $x' \in D$ such that $\bs{R}_{t}(x')+C \subsetneq \bs{R}_{t}(x) + C$.
\end{defn}

\begin{algorithm}[H]
\caption{Discarding Subroutine}\label{alg:discarding}
\begin{algorithmic}
\State \textbf{Compute:} $\mathcal{P}_{\text{pess},t}=\mathcal{P}^{(C)}_{\text{pess},t}(\mathcal{A}_t)$ \rev{as in Definition \ref{defn:pess}}
\For{$x \in \mathcal{S}_t \setminus \mathcal{P}_{\text{pess},t}$}
    \State \textbf{if} $\exists x' \in \mathcal{P}_{\text{pess},t} \ \forall \boldsymbol{y} \in \boldsymbol{R}_{t}(x):$  $\boldsymbol{R}_{t}(x') + \epsilon \boldsymbol{u}^* \subseteq \boldsymbol{y} + C$ \textbf{then}
    \State ~~\quad $\mathcal{S}_t = \mathcal{S}_t \setminus \{x\}$;
    \State \textbf{end if}
\EndFor
\end{algorithmic}
\end{algorithm}

\begin{remark}\label{rem:pess}
In the setting of Definition~\ref{defn:pess}, $\mathcal{P}^{(C)}_{\text{pess},t}(D)$ can be seen as the set of all nodes that yield maximal elements of a certain partial order relation on sets. Indeed, for $A,A'\subseteq \mathbb{R}^M$, let us write $A'\preceq_C A$ if and only if $A'\subseteq A+C$, which holds if and only if $A' + C\subseteq A+C$. Then, it is easy to check that the set relation $\preceq_C$ is a preorder, i.e., a reflexive and transitive binary relation, on the power set of $\mathbb{R}^M$. Let $\sim_C$ denote the symmetric part of this relation, i.e., $A'\sim_C A$ if and only if $A'\preceq_C A$ and $A\preceq_C A'$, which holds if and only if $A'+C=A+C$. Then, $\sim_C$ is an equivalence relation on the power set of $\mathbb{R}^M$ and the set of equivalence classes can be identified by the collection $\mathcal{M}_C:=\{A\subseteq\mathbb{R}^M\mid A=A+C\}$. Then, $\mathcal{M}_C$ is a partially ordered set with respect to $\preceq_C$ and the relation $\preceq_C$ coincides with the usual $\subseteq$ relation on $\mathcal{M}_C$. It then follows that $\mathcal{P}^{(C)}_{\text{pess},t}(D)$ is the set of all nodes $x\in D$ for which $\bs{R}_t(x)+C$ is a maximal element of $\{\bs{R}_t(x')+C\mid x'\in\mathcal{A}_t\}$ with respect to $\preceq_C$. In particular, since every finite nonempty partially ordered set has a maximal element, we have $\mathcal{P}^{(C)}_{\text{pess},t}(D)\neq\emptyset$. The reader is referred to \citet[Section~2.1]{setoptsurvey} for more details on set relations.
\end{remark}

VOGP computes pessimistic Pareto set of $\mathcal{A}_t$ by checking for each design in $\mathcal{A}_t$ whether there is another design that prevents it from being in the pessimistic Pareto set. To see if design $x$ prevents design $x^\prime$ from being in the pessimistic Pareto set, by solving a convex optimization problem, VOGP checks if every vertex of $\bs{R}_t(x^\prime)$ can be expressed as the sum of a vector in $C$ and a vector in $\bs{R}_t(x)$. VOGP refines the undecided set by utilizing the pessimistic Pareto set. It eliminates designs based on the following criteria: for a given design $x$, if all values inside its confidence hyperrectangle, when added to an "accuracy vector" $\epsilon\bs{u}^*$, are dominated with respect to $C$ by all values in the confidence hyperrectangle of another design $x'$, then the dominated design $x$ is discarded. The dominating design $x'$ should be a member of the pessimistic Pareto set.

\begin{defn}\label{defn:accuracy}
Let $A_C\coloneqq \bigcap_{\bs{u} \in \mathbb{S}^{M-1}}(\bs{u}+C)$ and $d_C\coloneqq\inf\{\|\bs{z}\|_2  \mid \bs{z} \in A_C\}$. Then, there exists a unique vector $\bs{z}^\ast\in A_C$ such that $d_C=\|\bs{z}^\ast\|_2$ and we define $\bs{u}^*\coloneqq \frac{\bs{z}^\ast}{d_C}$. 
\end{defn}

\begin{lem}\label{lem:ter}
Let $\bs{y},\bs{z}\in\mathbb{R}^M$ and $\bs{\Tilde{p}} \in \mathbb{B}(\frac{\epsilon}{d_C})$.
\begin{enumerate}[(i)]
\item If $\bs{y}+\bs{\Tilde{p}} \preccurlyeq_C \bs{z}$, then $\bs{y} \preccurlyeq_C \bs{z} + \epsilon\bs{u}^*$.
\item If $\bs{\Tilde{p}} \in \Int (\mathbb{B}(\frac{\epsilon} {d_C}))$ and $\bs{y}+\bs{\Tilde{p}} \preccurlyeq_C \bs{z}$, then $\bs{y} \prec_C \bs{z} + \epsilon\bs{u}^*$.
\end{enumerate}
\end{lem}

Due to its technical nature, the proof of Lemma \ref{lem:ter} is given in Section \ref{sec:thm1proof} along with the proofs of several other technical lemmas.

\begin{remark}
    $d_C$ corresponds to the magnitude of the smallest translation of a unit sphere so that the whole unit sphere is inside the cone $C$ (i.e., so that the origin dominates the unit sphere). $\bs{u}^*$ is the direction of this smallest translation. $d_C$ quantifies how hard it is to reach to a domination relation when using a specific ordering cone. We call this term the \textit{ordering hardness} of cone $C$. 
\end{remark}

\subsubsection{Computation of the pessimistic Pareto set}

Let $t\in\mathbb{N}$. According to Definition~\ref{defn:pess}, to check if a design $x\in\mathcal{A}_t$ is included in $\mathcal{P}_{\text{pess},t}$, the condition $\bs{R}_{t}(x') + C\subsetneq \bs{R}_{t}(x) + C$ needs to be checked for every $x'\in\mathcal{A}_t$. We first formulate this condition in terms of convex feasibility problems. These problems can also be expressed as convex optimization problems whose objective functions are zero and we solve them numerically using the CVXPY library \citep{agrawal2018rewriting,diamond2016cvxpy}.

Let $x,x'\in\mathcal{A}_t$. Since $\bs{R}_t(x')$ is a convex polytope, we have
\begin{align*}
\bs{R}_{t}(x')+C \subseteq \bs{R}_{t}(x) + C&\iff \bs{R}_{t}(x')\subseteq \bs{R}_{t}(x) + C\\
&\iff \forall \bs{v}'\in \vertex(\bs{R}_t(x'))\colon \bs{v}'\in \bs{R}_{t}(x) + C\\
&\iff \forall \bs{v}'\in \vertex(\bs{R}_t(x'))\ \exists \bs{y}\in \bs{R}_t(x)\colon \bs{y}\preccurlyeq_C \bs{v}'.
\end{align*}
Hence, for each $\bs{v}'\in\vertex(\bs{R}_t(x'))$, we check the condition $\exists \bs{y}\in \bs{R}_t(x)\colon \bs{y}\preccurlyeq_C \bs{v}'$ by solving the following convex feasibility problem:

\begin{equation}\tag{$F_t(x,\bs{v}')$}\label{Feas1}
\begin{aligned}
& \underset{\bs{y}}{\text{minimize}}
& & 0 \\
& \text{subject to}
& & \bs{y} \in \mathbb{R}^M, \\
&&& y^j \geq \mu^j_s(x) - \beta_s^{1/2} \sigma^j_s(x), \quad \forall j \in [M], \forall s \in [t] \\
&&& y^j \leq \mu^j_s(x) + \beta_s^{1/2} \sigma^j_s(x), \quad \forall j \in [M], \forall s \in [t] \\
&&& \bs{w}_n^{\mathsf{T}}\bs{y}\leq \bs{w}_n^{\mathsf{T}}\bs{v}', \quad \forall n \in [N].
\end{aligned}
\end{equation}
Moreover, under the condition $\bs{R}_{t}(x')+C \subseteq \bs{R}_{t}(x) + C$, we have
\begin{align*}
\bs{R}_{t}(x')+C \neq \bs{R}_{t}(x) + C&\iff \bs{R}_{t}(x)\not\subseteq \bs{R}_{t}(x') + C\\
&\iff \exists \bs{v}\in \vertex(\bs{R}_t(x))\colon \bs{v}\notin \bs{R}_{t}(x') + C\\
&\iff \exists \bs{v}\in \vertex(\bs{R}_t(x)) \not\exists \bs{y}'\in \bs{R}_t(x')\colon \bs{y}'\preccurlyeq_C \bs{v}.
\end{align*}
Hence, for each $\bs{v}\in \vertex(\bs{R}_t(x))$, we check the condition $\exists \bs{y}'\in \bs{R}_t(x')\colon \bs{y}'\preccurlyeq_C \bs{v}$ by solving \eqref{Feas2} defined in the same way:
\begin{equation}\tag{$F_t(x',\bs{v})$}\label{Feas2}
\begin{aligned}
& \underset{\bs{y}'}{\text{minimize}}
& & 0 \\
& \text{subject to}
& & \bs{y}' \in \mathbb{R}^M, \\
&&& (y')^j \geq \mu^j_s(x') - \beta_s^{1/2} \sigma^j_s(x'), \quad \forall j \in [M], \forall s \in [t] \\
&&& (y')^j \leq \mu^j_s(x') + \beta_s^{1/2} \sigma^j_s(x'), \quad \forall j \in [M], \forall s \in [t] \\
&&& \bs{w}_n^{\mathsf{T}}\bs{y}'\leq \bs{w}_n^{\mathsf{T}}\bs{v}, \quad \forall n \in [N].
\end{aligned}
\end{equation}
Hence, if there exists $x'\in\mathcal{A}_t$ such that \eqref{Feas1} is feasible for every $\bs{v}'\in \vertex(\bs{R}_t(x'))$ and \eqref{Feas2} is infeasible for at least one $\bs{v}\in\vertex(\bs{R}_t(x))$, then $x$ is not added to $\mathcal{P}_{\text{pess},t}$. Otherwise, $x$ is added to $\mathcal{P}_{\text{pess},t}$.

\subsubsection{Computation of discarded designs} 

In its original form, the discarding criteria of VOGP (see Algorithm~\ref{alg:discarding}) requires checking domination relations between infinitely many point pairs from two confidence hyperrectangles. In this section, we show that the sufficient number of checks can be reduced significantly by considering only the vertex pairs of the hyperrectangles, allowing for an efficient implementation. This is achieved thanks to the next lemma. For a nonempty convex polytope $A\subseteq \mathbb{R}^M$, we denote by $\vertex(A)$ its set of all vertices.

\begin{lem} \label{lem:eqdisc1}
    Let $x,x'\in\mathcal{X}$ and $t\in\mathbb{N}$. For the cumulative confidence hyperrectangles $\bs{R}_t(x)$, $\bs{R}_t(x')$, the following statements are equivalent.
    \begin{itemize}
    \item[(i)] $\forall  \bs{y} \in \bs{R}_{t}(x)\ \forall \bs{y}' \in \bs{R}_{t}(x')\colon \bs{y} \preccurlyeq_C \bs{y}'$.
    \item[(ii)] $\forall \bs{v}\in \vertex(\bs{R}_{t}(x))\ \forall \bs{v}'\in \vertex(\bs{R}_{t}(x'))\colon \bs{v} \preccurlyeq_C \bs{v}'$.
\end{itemize}
\end{lem}

\noindent \textit{Proof.}
It is clear that (i) implies (ii) since $\vertex(\bs{R}_t(x))\subseteq \bs{R}_t(x)$ and $\vertex(\bs{R}_t(x'))\subseteq \bs{R}_t(x')$. To show the reverse implication, suppose that (ii) holds; let $\bs{y}\in \bs{R}_t(x)$ and $\bs{y}'\in\bs{R}_t(x')$. We proceed in two steps.

First, let us further assume that $\bs{y}\in \vertex(\bs{R}_t(x))$. Moreover, since $\bs{y}'\in\bs{R}_t(x')$ and $\bs{R}_t(x')$ is a convex polytope, we may write
\[
\bs{y}'=\sum_{\bs{v}'\in \vertex(\bs{R}_t(x'))} \lambda_{\bs{v}'}\bs{v}',
\]
where $\lambda_{\bs{v}'}\geq 0$ for each $\bs{v}'\in\vertex(\bs{R}_t(x'))$ such that
\[
\sum_{\bs{v}'\in \vertex(\bs{R}_t(x'))} \lambda_{\bs{v}'}=1.
\]
Note that (ii) implies that
\[
\forall \bs{v}'\in\vertex(\bs{R}_t(x'))\colon \bs{y}\preccurlyeq_C\bs{v}',
\]
which implies that
\[
\forall \bs{v}'\in\vertex(\bs{R}_t(x'))\colon \lambda_{\bs{v}'}\bs{y}\preccurlyeq_C\lambda_{\bs{v}'}\bs{v}'.
\]
Then, summing over all $\bs{v}'\in\vertex(\bs{R}_t(x'))$ yields
\[
\bs{y}=\sum_{\bs{v}'\in\vertex(\bs{R}_t(x'))}\lambda_{\bs{v}'}\bs{y}\preccurlyeq_C\sum_{\bs{v}'\in\vertex(\bs{R}_t(x'))}\lambda_{\bs{v}'}\bs{v}'=\bs{y}'.
\]
Hence, (i) holds in this step.

Next, we consider the general case, where $\bs{y}\in \bs{R}_t(x)$. Since $\bs{R}_t(x)$ is a convex polytope, we may write
\[
\bs{y}=\sum_{\bs{v}\in \vertex(\bs{R}_t(x))} \alpha_{\bs{v}}\bs{v},
\]
where $\alpha_{\bs{v}}\geq 0$ for each $\bs{v}\in\vertex(\bs{R}_t(x))$ such that
\[
\sum_{\bs{v}\in \vertex(\bs{R}_t(x))} \alpha_{\bs{v}}=1.
\]
By the first step, we have
\[
\forall \bs{v}\in\vertex(\bs{R}_t(x))\colon \bs{v}\preccurlyeq_C \bs{y}'.
\]
Then, we obtain
\[
\forall \bs{v}\in\vertex(\bs{R}_t(x))\colon \alpha_{\bs{v}}\bs{v}\preccurlyeq_C \alpha_{\bs{v}}\bs{y}'.
\]
Finally, summing over all $\bs{v}\in\vertex(\bs{R}_t(x))$ yields
\[
\bs{y}=\sum_{\bs{v}\in\vertex(\bs{R}_t(x))}\alpha_{\bs{v}}\bs{v}\preccurlyeq_C\sum_{\bs{v}\in\vertex(\bs{R}_t(x))}\alpha_{\bs{v}}\bs{y}'=\bs{y}'.
\]
Hence, (i) holds in the general case too.
\qed 

Let $t\in\mathbb{N}$. In Algorithm~\ref{alg:discarding}, to decide if a design $x\in\mathcal{S}_t$ is going to be discarded, the condition
\[
\exists x' \in \mathcal{P}_{\text{pess},t} \ \forall \bs{y} \in \bs{R}_{t}(x): \bs{R}_{t}(x') + \epsilon \bs{u}^* \subseteq \bs{y} + C
\]
needs to be checked, which requires enumerating all designs in $\mathcal{P}_{\text{pess},t}$. Given $x'\in \mathcal{P}_{\text{pess},t}$, observe that
\begin{align*}
    \forall \bs{y} \in \bs{R}_{t}(x)\colon \bs{R}_{t}(x') + \epsilon \bs{u}^* \subseteq \bs{y} + C &\iff \forall \bs{y} \in \bs{R}_{t}(x)\ \forall \bs{y}' \in \bs{R}_{t}(x')\colon \bs{y} \preccurlyeq_C \bs{y}' + \epsilon \bs{u}^* \\
    &\iff \forall \bs{y} \in \bs{R}_{t}(x)\ \forall \bar{\bs{y}} \in \bs{R}_{t}(x')+\epsilon \bs{u}^* \colon \bar{\bs{y}} \preccurlyeq_C \bs{y}~.
\end{align*}
Hence, the last condition can be checked equivalently instead of the original condition. Notice that $\bs{R}_{t}(x')+\epsilon \bs{u}^* $ is a shifted hyperrectangle, which is also a hyperrectangle. Therefore, by Lemma \ref{lem:eqdisc1}, the above conditions are also equivalent to the following:
\begin{equation}
    \forall \bs{v}\in \vertex(\bs{R}_{t}(x)), \forall \bs{v}'\in \vertex(\bs{R}_{t}(x'))\colon \bs{v} \preccurlyeq_C \bs{v}'+\epsilon \bs{u}^*~. \label{eq:fast}
\end{equation}
Hence, in the implementation of VOGP, to determine if $x\in \mathcal{S}_t$ is going to be discarded, we check if there exists $x' \in \mathcal{P}_{\text{pess},t}$ such that Equation \eqref{eq:fast} holds.

\subsection{Pareto identification} VOGP aims to identify the designs that are not dominated by any other design with high probability with respect to the ordering cone $C$. It does so by pinpointing designs that, after adding the accuracy vector to the values in their confidence hyperrectangles, remain non-dominated with respect to $C$ when compared to the values in the confidence hyperrectange of any other design. The identified designs are moved from the set of undecided designs to the predicted Pareto set. It is important to note that, once a design becomes a member of the predicted Pareto set, it remains a permanent member of that set. The illustration of Pareto identification phase can be seen in Figure~\ref{fig:enter-label} (Right).

\begin{algorithm}[H]
\caption{Pareto Identification Subroutine}\label{alg:pareto}
\begin{algorithmic}
\State \rev{\textbf{Compute:} Set of active designs $\mathcal{W}_t = \mathcal{S}_t \cup \mathcal{P}_t$};
\For{$x \in \mathcal{S}_t$}
    \If{$\nexists x' \in \mathcal{W}_t\colon (\boldsymbol{R}_{t}(x) + \epsilon \boldsymbol{u}^* + C) \cap \boldsymbol{R}_{t}(x') \neq \emptyset$}
    \State ~~\quad $\mathcal{S}_t = \mathcal{S}_t \setminus \{x\}$, $\mathcal{P}_t = \mathcal{P}_t \cup \{x\}$;
    \EndIf
\EndFor
\end{algorithmic}
\end{algorithm}

\subsubsection{Computation of the Pareto identification subroutine}

Let $t\in\mathbb{N}$. According to Algorithm~\ref{alg:pareto}, to determine if a design $x\in\mathcal{S}_t$ is going to be added to $\mathcal{P}_t$, the condition
\[
(\bs{R}_{t}(x) + \epsilon \bs{u}^* + C) \cap \bs{R}_{t}(x') \neq \emptyset
\]
needs to be checked for every $x'\in\mathcal{W}_t$, which can be rewritten as
\begin{equation}\label{eq:epseq}
    \exists \bs{y}' \in \bs{R}_{t}(x')\ \exists \bs{y} \in \bs{R}_{t}(x): \bs{y} + \epsilon \bs{u}^* \preccurlyeq_C \bs{y}' ~. 
\end{equation}
We first restate \eqref{eq:epseq} as a convex feasibility problem. Then, in the implementation of VOGP, this convex feasibility problem is solved using the CVXPY library \citep{agrawal2018rewriting,diamond2016cvxpy}. 

Note that \eqref{eq:epseq} can be checked by solving the following feasibility problem:
\begin{equation*}
\begin{aligned}
& \underset{\bs{y}, \bs{y}'}{\text{minimize}}
& & 0 \\
& \text{subject to}
& & \bs{y}, \bs{y}' \in \mathbb{R}^M, \\
&&& y^j \geq \mu^j_s(x) - \beta_s^{1/2} \sigma^j_s(x), \quad \forall j \in [M], \forall s \in [t] \\
&&& y^j \leq \mu^j_s(x) + \beta_s^{1/2} \sigma^j_s(x), \quad \forall j \in [M], \forall s \in [t] \\
&&& (y')^j \geq \mu^j_s(x') - \beta_s^{1/2} \sigma^j_s(x'), \quad \forall j \in [M], \forall s \in [t] \\
&&& (y')^j \leq \mu^j_s(x') + \beta_s^{1/2} \sigma^j_s(x'), \quad \forall j \in [M], \forall s \in [t] \\
&&& \bs{w}_n^{\mathsf{T}} (\bs{y}'-\bs{y} - \epsilon \bs{u}^*) \geq 0, \quad \forall n \in [N]~.
\end{aligned}
\end{equation*}
If this problem turns out to be infeasible for all $x' \in \mathcal{W}_t$, then $x$ is added to $\mathcal{P}_t$. Otherwise, $x$ is not added to $\mathcal{P}_t$.

\subsection{Evaluating} VOGP selects the design whose confidence hyperrectangle has the widest diagonal. The diagonal of the hyperrectangle $\bs{R}_{t}(x)$ is given by $\omega_t(x) = \max _{\bs{y}, \bs{y}^{\prime} \in \bs{R}_{t}(x)}\left\|\bs{y}-\bs{y}^{\prime}\right\|_2^2$. The motivation behind this step is to acquire as much information about the objective space as possible, so that the distinction between Pareto and non-Pareto designs can be made fast with high probability.

\begin{algorithm}[H]
\caption{Evaluating Subroutine}\label{alg:evaluating}
\begin{algorithmic}
\If{$\mathcal{S}_t \neq \emptyset$}
    \State Select design $x_t \in \underset{x \in \mathcal{W}_t}{\arg\max}~ \omega_t(x)$;
    \State Observe $\boldsymbol{y}_t = \boldsymbol{f}(x_t) + \boldsymbol{\nu}_t$;
\EndIf
\end{algorithmic}
\end{algorithm}

\section{Main theoretical results and analysis}
In this section, we provide theoretical guarantees for VOGP. We first state the main theorems.

\begin{theorem}\label{thm:main}
    Let $\eta := \sigma^{-2} / \ln (1+\sigma^{-2})$. When VOGP is run, at least with probability $1 - \delta$, an $(\epsilon, \delta)$-Pareto set can be identified with no more than $T$ function evaluations, where
\begin{equation}
     T \coloneqq \min \Big\{ t \in \mathbb{N} :  \sqrt{\frac{8\beta_{t}  \sigma^2 \eta M \gamma_{t} }{t}}  < \frac{\epsilon}{d_C} \Big\} ~. \label{eq:tg}
\end{equation}
\end{theorem}

This theorem provides an implicit bound on the sample complexity in terms of the maximum information gain $\gamma_t$. Given a specific kernel, one can numerically solve \eqref{eq:tg} to compute $T$. It is obvious that when the slackness term $\epsilon$ gets smaller, sample complexity increases. Similarly, as the cone-specific ordering hardness $d_C$ gets larger, sample complexity increases. 

Here, we sketch the idea behind the proof of Theorem \ref{thm:main}. We first define $E$ to be the event that the confidence hyperrectangles of designs include their true objective values and prove that $\mathbb{P}(E)\geq 1-\delta$. Then, we define $d_C(\cdot)$ and $A_C(\cdot)$ and show that they have the homogeneity property (see Lemma~\ref{lem:scalability}). Using this result, we establish the stopping criterion for VOGP by demonstrating that the algorithm concludes when the diagonals of the confidence hyperrectangles are less than $\frac{\epsilon}{d_C}$ (see Lemmata~\ref{lem:ustar}, \ref{lem:terminationcondition_withdirection}). Finally, by bounding the diagonals using the maximum information gain and noting their monotonic decrease, we determine the sample complexity. The resulting sample complexity is the smallest number of samples needed for the upper bound of the diagonals to drop below $\frac{\epsilon}{d_C}$ (see Lemma~\ref{lem:infogap}).

The next theorem provides a more explicit bound on the sample complexity for a specific choice of the kernel. 

\begin{theorem}\label{thm:kernel}
    Let the GP kernel $\bs{k}$ have the multiplicative form $\bs{k}(x, x^{\prime})=[\tilde{k}(x, x^{\prime}) k^*(p,q)]_{p, q \in[M]}$, where $\tilde{k}\colon\mathcal{X}\times\mathcal{X}\to\mathbb{R}$ is a kernel for the design space and $k^*\colon[M]\times[M]\to\mathbb{R}$ is a kernel for the objective space. Assume that $\tilde{k}$ is a squared exponential kernel. Then, with at least $1 - \delta$ probability, the sample complexity of VOGP is given by $T = \tilde{\mathcal{O}}\left( \frac{ d_C^2}{\epsilon^2} \right)$.
\end{theorem}

To prove Theorem~\ref{thm:kernel}, we use the bound on maximum information gain for single output GPs given in \cite{vakilibounds} to bound the maximum information gain of an $M$-output GP (see Lemma~\ref{lem:siribound}). Then,
by algebraic manipulations, we establish that the proposed sampling budget satisfies the inequality in \eqref{eq:tg}. The sample complexity of VOGP when using Matérn kernel was also established using a similar approach and can be found in Appendix \ref{sec:mat}. Theorems~\ref{thm:main} and \ref{thm:kernel} provide sample complexity analysis of VOGP in terms of ordering hardness $d_C$. 

Theorem~\ref{thm:main} provides a bound that is applicable in a very general context where the kernel $\bs{k}$ can be any bounded positive definite kernel. Theorem~\ref{thm:kernel} provides a more explicit and practical formulation of sample complexity. The following remark compares the established bounds with the bounds in prior work.

\begin{remark}
The sample complexity bounds of VOGP established in Theorems~\ref{thm:main} and \ref{thm:kernel} exactly match the sample complexity bounds of previous works in multi-objective Bayesian optimization literature, namely PAL \citep{zuluaga2013active}, $\epsilon$-PAL \citep{JMLR:v17:15-047} in terms of $\epsilon$. These works present bounds of $\mathcal{O}(\frac{1}{\epsilon^{2+\rho}})$ for any $\rho > 0$. Additionally, some studies focusing on Pareto identification in the multi-objective multi-armed bandit setting, which do not utilize GPs, also provide sample complexity bounds \citep{ararat2023vector,drugan2017pac,auer2016pareto} that matches those of VOGP up to logarithmic terms with $\tilde{\mathcal{O}}(\frac{1}{\epsilon^2})$. In contrast, other baseline methods in this study that employ GPs—such as JES, MESMO, and EHVI— do not provide sample complexity bounds.
\end{remark}

The following two subsections are dedicated to the proofs of Theorems~\ref{thm:main} and \ref{thm:kernel}.

\subsection{Proof of Theorem \ref{thm:main}} \label{sec:thm1proof}

Let \rev{accuracy parameter} $\epsilon>0$ and \rev{confidence parameter} $\delta\in(0,1)$ be given. Let $\mathbb{N}\coloneqq\{1,2,\ldots\}$.

\begin{lem}\label{lem:prob} Let us define the event \rev{that confidence hyperrectangles include the true objective values of designs as}
	\begin{align*}
	E \coloneqq \left\{ \forall j \in [M]\ \forall t\in\mathbb{N}\ \forall x \in \mathcal{X}\colon |f^j(x) - \mu^j_{t }(x) | \leq \beta^{1/2}_t \sigma^j_{t}(x) \right\},
	\end{align*}
	where
 \[
 \beta_t \coloneqq \ln (\frac{M\pi^2 |\mathcal{X}| t^2}{3\delta }).
 \]
 Then, $\mathbb{P}(E)\geq 1-\delta$.
\end{lem}

\textit{Proof.} For an event $E^\prime$, we denote by $\mathbb{I}(E^\prime)$ its probabilistic indicator function, i.e., $\{\mathbb{I}(E^\prime)=1\}=E^{\prime}$ and $\{\mathbb{I}(E^\prime)=0\}=\Omega\setminus E^{\prime}$, where $\Omega$ is the underlying sample space. Note that
\begin{align}
1 - \mathbb{P}(E)  &= \mathbb{E}\left[ \mathbb{I}  \left(\left\{ \exists j\in[M]\ \exists t \in\mathbb{N}\ \exists x \in \mathcal{X} \colon |f^j(x) - \mu^j_{t }(x) | > \beta^{1/2}_t \sigma^j_{t} (x) \right\}\right) \right] \notag \\
& \leq \mathbb{E} \left[ \sum^M_{j=1}\sum_{t=1}^\infty \sum_{x \in \mathcal{X} }  \mathbb{I} \left(\left\{ | f^j(x) - \mu^j_{t }(x) | > \beta^{1/2}_t \sigma^j_{t } (x)\right\} \right) \right] \notag \\
& = \sum^M_{j=1}\sum_{t=1}^\infty  \sum_{x \in \mathcal{X} }  \mathbb{E} \left[  \mathbb{E} \left[  \mathbb{I} \left(\left\{ | f^j(x) - \mu^j_{t }(x) | > \beta^{1/2}_t  \sigma^j_{t}(x) \right\}\right) \big\vert \bs{y}_{[t-1]}  \right]  \right]\label{eq520}\\   
& = \sum^M_{j=1}\sum_{t =1}^\infty \sum_{x \in \mathcal{X} }  \mathbb{E} \left[  \mathbb{P} \left( \left\{ | f^j(x) - \mu^j_{t }(x) | > \beta^{1/2}_t  \sigma^j_{t } (x) \right\} \big\vert \bs{y}_{[t-1]} \right)   \right] \notag \\ 
& \leq  \sum^M_{j=1}\sum_{t = 1}^\infty \sum_{x \in \mathcal{X} }  2 e^{- \beta_t /2}  \label{eq5204}\\
& =  2M |\mathcal{X}|\sum_{t = 1}^\infty  e^{- \beta_t  /2}\notag\\
& =  2M|\mathcal{X}| \sum_{t = 1}^\infty  \left(\frac{M\pi^2 |\mathcal{X}| {t^2}}{3\delta }\right)^{-1} \notag \\
& = \frac{6\delta}{\pi^2} \sum_{t =1 }^\infty \frac{1}{t^{2}} =\delta ~, \notag
\end{align}
where (\ref{eq520}) uses the tower rule and linearity of expectation and (\ref{eq5204}) uses Gaussian tail bound; here note that, given $\bs{y}_{[t-1]}$, the conditional distribution of $f^j(x)$ is $\mathcal{N}(\mu^j_{t}(x),\sigma^j_{t}(x))$.
\qed

\rev{For each $s>0$, let us introduce the set $A_C(s)$, which consists of vectors that dominate all points on a sphere of radius $s$ centered at the origin. We define $d_C(s)$ as the infimum of the norms of the elements in this set. These concepts are formally expressed as follows:}
\[
A_C(s) \coloneqq \bigcap_{\bs{u} \in \mathbb{S}^{M-1}}(s\bs{u}+C),\quad d_C(s)\coloneqq\inf\{\|\bs{z}\|_2 \mid \bs{z} \in A_C(s)\}.
\]

 When $s=1$, we recover the quantities in Definition \ref{defn:accuracy}, since $d_C(1)= d_C$, $A_C(1)= A_C$.

\begin{lem}\label{lem:scalability}
Let $s>0$. Then, 
\[
A_C(s)=\left\{\bs{z} \in \mathbb{R}^M \mid \bs{w}_n^{\top} \bs{z} \geq s,  ~\forall n \in[N]\right\}~.
\]
In particular, $A_C(s) = sA_C(1)\coloneqq \{s\bs{y} \mid \bs{y} \in A_C(1)\}$ and $d_C(s)=s d_C(1)$. Moreover, there exists a unique $\bs{z}^s\in A_C(s)$ such that $d_C(s)=\|\bs{z}^s\|_2$\revf{, where the superscript denotes dependence on $s$ (not exponentiation)}. 
\end{lem}

\textit{Proof.}
Let $s>0$. Then, \rev{it can be shown that}
\begin{align}
\bs{y} \in A_C(s) &\Leftrightarrow \bs{y} \in s \bs{u} + C, ~\forall \bs{u} \in \mathbb{S}^{M-1} \notag \\
						   &\Leftrightarrow \bs{y} - s \bs{u} \in C, ~\forall \bs{u} \in \mathbb{S}^{M-1} \notag\\
						   &\Leftrightarrow \bs{w}_n^{\top} ( \bs{y} - s \bs{u} ) \geq 0, ~\forall n \in [N], ~\forall \bs{u} \in \mathbb{S}^{M-1}\label{eq:conedom1}\\
 & \Leftrightarrow 
\bs{w}_n^{\top} \bs{y} \geq  s \bs{w}_n^{\top} \bs{u}, ~\forall n \in [N], ~\forall \bs{u} \in \mathbb{S}^{M-1} \notag\\ 
& \Leftrightarrow \bs{w}_n^{\top} \bs{y} \geq s \sup _{\bs{u} \in \mathbb{S}^{M-1}}\left(\bs{w}_n^{\top} \bs{u} \right), ~\forall n \in [N]  ~,\label{eq:567}
\end{align}

\rev{where  (\ref{eq:conedom1}) follows from (\ref{eq:conedomination}).} Let $n\in[N]$. By the definition of $\bs{w}_n$, we have
\begin{align}
    \sup_{\bs{u} \in \mathbb{S}^{M-1}} \bs{w}_n^{\top} \bs{u} = \sup_{\bs{u} \in \mathbb{B}(1)} \bs{w}_n^{\top} \bs{u} = || \bs{w}_n ||_2 =1 ~. \label{eq:wn}
\end{align}
Combining (\ref{eq:567}) and (\ref{eq:wn}), we get
\[
    \bs{y} \in A_C(s) \Leftrightarrow \bs{w}_n^{\top} \bs{y} \geq s ,~\forall n \in [N] ~. 
\]
Therefore, $A_C(s)=\left\{\bs{z} \in \mathbb{R}^M \mid \bs{w}_n^{\top} \bs{z} \geq s,  ~\forall n \in[N]\right\}$, which implies that $A_C(s)=s A_C(1)$ and
hence $d_C(s)=sd_C(1)$.

The existence and uniqueness of $\bs{z}^s$ is a direct consequence of the strict convexity and continuity of the $\ell_2$-norm $\|\cdot\|_2$ together with the closedness and convexity of $A_C(s)$.\qed 

\begin{remark}
The last part of Lemma~\ref{lem:scalability} justifies the definition of $\bs{u}^\ast$ (see Definition~\ref{defn:accuracy}).
\end{remark}

In what follows, we denote by $\mathbb{S}^{M-1}(r)$ the boundary of $\mathbb{B}(r)$, where $r>0$; in particular, $\mathbb{S}^{M-1}(1)=\mathbb{S}^{M-1}$ is the unit sphere. We next prove Lemma~\ref{lem:ter} as a consequence of Lemma~\ref{lem:scalability}. 

\textit{Proof of Lemma~\ref{lem:ter}}.
\begin{enumerate}[(i)]
\item Suppose that $\bs{y}+\bs{\tilde{p}}\preccurlyeq_C\bs{z}$. Equivalently, we have $\bs{z}\in\bs{y}+\bs{\tilde{p}}+C$. \rev{Then, by (\ref{eq:conedomination}), we have}
\begin{equation}\label{eq:w1}
\forall n\in[N]\colon \bs{w}_n^{\mathsf{T}}(\bs{z}-\bs{y}-\bs{\tilde{p}})\geq 0.
\end{equation}
Since
$\bs{\tilde{p}}\in  \mathbb{B}(\frac{\epsilon}{d_C})$, by Cauchy-Schwarz inequality, we have 
\begin{equation}\label{eq:w2}
\abs{\bs{w}_n^{\mathsf{T}}\bs{\tilde{p}}}\leq \norm{\bs{w}_n}_2\norm{\bs{\tilde{p}}}_2\leq \frac{\epsilon}{d_C}
\end{equation}
for every $n\in[N]$. Moreover, by Lemma~\ref{lem:scalability} \rev{and Definition \ref{defn:accuracy}}, we have
\begin{equation}\label{eq:w3}
\bs{w}_n^{\mathsf{T}}\bs{u}^*=\frac{\bs{w}_n^{\mathsf{T}}\bs{z}^*}{d_C}\geq \frac{1}{d_C}
\end{equation}
for every $n\in[N]$. \rev{Note that $\bs{w}_n^{\mathsf{T}}\bs{z}^* \geq 1$ follows from the definition of $A_C$ and that $\bs{z}^* \in A_C$.}
Hence, by \eqref{eq:w1}, \eqref{eq:w2}, and \eqref{eq:w3}, we get
\[
\bs{w}_n^{\mathsf{T}}(\bs{z}+\epsilon\bs{u}^*-\bs{y})\geq \bs{w}_n^{\mathsf{T}}\bs{\tilde{p}}+\epsilon \bs{w}_n^{\mathsf{T}}\bs{u}^*\geq -\frac{\epsilon}{d_C}+\frac{\epsilon}{d_C}\geq 0
\]
for every $n\in[N]$. This shows that $\bs{z}+\epsilon\bs{u}^*-\bs{y}\in C$, i.e., $\bs{y} \preccurlyeq_C \bs{z} + \epsilon\bs{u}^*$.
\item Suppose that $\bs{\Tilde{p}} \in \Int (\mathbb{B}(\frac{\epsilon} {d_C}))$ and $\bs{y}+\bs{\tilde{p}}\preccurlyeq_C\bs{z}$. In particular, \eqref{eq:w1} and \eqref{eq:w3} still hold. Since $\bs{\Tilde{p}} \in \Int (\mathbb{B}(\frac{\epsilon} {d_C}))$, by Cauchy-Schwarz inequality, we have 
\begin{equation}\label{eq:w4}
\abs{\bs{w}_n^{\mathsf{T}}\bs{\tilde{p}}}\leq \norm{\bs{w}_n}_2\norm{\bs{\tilde{p}}}_2< \frac{\epsilon}{d_C}
\end{equation}
for every $n\in[N]$. Hence, by \eqref{eq:w1}, \eqref{eq:w3}, and \eqref{eq:w4}, we get 
\[
\bs{w}_n^{\mathsf{T}}(\bs{z}+\epsilon\bs{u}^*-\bs{y})\geq \bs{w}_n^{\mathsf{T}}\bs{\tilde{p}}+\epsilon \bs{w}_n^{\mathsf{T}}\bs{u}^*> -\frac{\epsilon}{d_C}+\frac{\epsilon}{d_C}\geq 0
\]
for every $n\in[N]$. This shows that $\bs{z}+\epsilon\bs{u}^*-\bs{y}\in \Int(C)$, i.e., $\bs{y} \prec_C \bs{z} + \epsilon\bs{u}^*$.
\end{enumerate}
\qed

\rev{In the following lemma, we show that the predicted Pareto set returned by VOGP satisfies the first success condition. The proof establishes that under event $E$, every Pareto optimal design $x \in P^*$ has a corresponding design $z \in \hat{\mathcal{P}}$ such that $\bs{f}(x) \preccurlyeq_C \bs{f}(z) + \epsilon \bs{u}^*$. If $x$ is not directly in $\hat{\mathcal{P}}$, it must have been discarded at some round. In that case, the discarding subroutine ensures the existence of a design $z_1$ in the pessimistic Pareto set that caused $x$ to be discarded. If $z_1$ is in $\hat{\mathcal{P}}$, the success condition holds. Otherwise, $z_1$ must have been excluded from the pessimistic Pareto set by another design $z_2$. If $z_2 \in \hat{\mathcal{P}}$, the success condition holds and if it does not, some design $z_3$ must have excluded $z_2$ from the pessimistic Pareto set, continuing this process iteratively. The chain eventually leads to a design that is included in $\hat{\mathcal{P}}$, ensuring that the success condition holds.}

\begin{lem}\label{lem:returnspareto}
    Under event $E$, the set $\hat{\mathcal{P}}$ returned by VOGP satisfies condition (i) in Definition~\ref{defn:success}.
\end{lem}

\textit{Proof.} We assume that event $E$ occurs and claim that for every $x \in P^*$, there exists $z \in \hat{\mathcal{P}}$ such that $\bs{f}(x) \preccurlyeq_C \bs{f}(z) + \epsilon \bs{u}^*$. If $x \in \hat{\mathcal{P}}$, then the claim holds with $z = x$, i.e., $\bs{f}(x) \preccurlyeq_C \bs{f}(x)+  \epsilon \bs{u}^*$, since $\epsilon  \bs{u}^*\in C$. Suppose that $x \not\in \hat{\mathcal{P}}$. Then, $x$ must have been discarded at some round $s_1$. By the discarding rule, there exists $z_1 \in \mathcal{P}_{\text{pess}, s_1}$ such that 
\begin{equation}\label{eq:discardingrule}
    \bs{R}_{s_1}(z_1)+\epsilon \bs{u}^*\subseteq \bs{y} + C
\end{equation}
holds for every $\bs{y} \in \bs{R}_{s_1}(x)$.

At each round,  the initial confidence hyperrectangle of an arbitrary design $x'\in\mathcal{X}$ \rev{given by (\ref{eq:hyperrec}) can be expressed as }
\begin{equation}
    \bs{Q}_t(x')=\left\{\bs{y}\in\mathbb{R}^M\mid \bs{\mu}_t(x')-\beta_t^{1 / 2} \bs{\sigma}_t(x') \preccurlyeq_{\mathbb{R}_+^M} \bs{y} \preccurlyeq_{\mathbb{R}_+^M} \bs{\mu}_t(x')+\beta_t^{1 / 2} \bs{\sigma}_t(x')\right\} ~ \label{eq:q}
\end{equation}
and the initial confidence hyperrectangle is intersected with previous hyperrectangles to obtain the confidence hyperrectangle of the current round, that is,
 \begin{equation}
     \bs{R}_t(x')=\bs{R}_{t-1}(x') \cap \bs{Q}_t(x') \label{eq:inter}~.
 \end{equation}
It can be checked that, due to Lemma \ref{lem:prob}, we have $\bs{f}(x') \in \bs{R}_{s_1}(x')$ for every $x'\in\mathcal{X}$ under event $E$. In particular, by \eqref{eq:discardingrule}, we have
\begin{equation}\label{def:discardingrule_supp2}
    \bs{R}_{s_1}(z_1)+\epsilon\bs{u}^* \subseteq \bs{f}(x) + C ~,
\end{equation}
where $z_1 \in \mathcal{P}_{\text{pess}, s_1}$.

Since $\bs{f}(z_1) \in \bs{R}_{s_1}(z_1)$, \eqref{def:discardingrule_supp2} implies that $\bs{f}(z_1) + \epsilon \bs{u}^*\in \bs{f}(x) + C$, which is equivalent to $\bs{f}(x) \preccurlyeq_C \bs{f}(z_1) + \epsilon \bs{u}^*$. Therefore, if $z_1 \in \hat{\mathcal{P}}$, then the claim holds by choosing $z = z_1$.

If $z_1 \not\in \hat{\mathcal{P}}$, then it must have been discarded at some round $s_2 \geq s_1$. Because VOGP discards from the set $\mathcal{S}_t \setminus \mathcal{P}_{\text{pess},t}$ at any round $t$, we have $z_1 \not\in \mathcal{P}_{\text{pess},s_2}$. Then, using the definition of the pessimistic Pareto set (see Definition~\ref{defn:pess}), there exists $z_2 \in \mathcal{A}_{s_2}$ such that $\bs{R}_{s_2}(z_2)+C \subsetneq \bs{R}_{s_2}(z_1) + C$, which implies that

\begin{equation}\label{def:pesssetule}
      \bs{R}_{s_2}(z_2) \subseteq \bs{R}_{s_2}(z_2)+C \subseteq\bs{R}_{s_2}(z_1) + C ~.
\end{equation}

To proceed, we use \eqref{def:discardingrule_supp2} and the fact that $C + C = C$ to obtain
\begin{equation}\label{eq:hhj}
\bs{R}_{s_1}(z_1)+\epsilon\bs{u}^*+C \subseteq \bs{f}(x) + C +C = \bs{f}(x) + C.
\end{equation}
In addition, \eqref{def:pesssetule} implies that
\[
\bs{R}_{s_2}(z_2)+\epsilon \bs{u}^*\subseteq \bs{R}_{s_1}(z_1)+\epsilon\bs{u}^*+C ~.
\]
Combining the above display with \eqref{eq:hhj} yields
\[
\bs{R}_{s_2}(z_2)+\epsilon \bs{u}^*\subseteq \bs{R}_{s_1}(z_1)+\epsilon\bs{u}^*+C\subseteq \bs{f}(x) + C ~.
\]
According to Lemma~\ref{lem:prob}, under event $E$, we have $\bs{f}(z_2) \in \bs{R}_{s_2}(z_2)$. Hence, it holds that
\[
\bs{f}(z_2) + \epsilon \bs{u}^*\in \bs{f}(x) + C ~.
\]
So, if $z_2 \in \hat{\mathcal{P}}$, then the claim holds with $ z = z_2$.

If $z_2 \not\in \hat{\mathcal{P}}$, then $z_2$ must have been discarded at some round $s_3 \geq s_2$. Because VOGP discards from the set $\mathcal{S}_t \setminus \mathcal{P}_{\text{pess},t}$ at any round $t$, we have $z_2 \not\in \mathcal{P}_{\text{pess},s_3}$. Then, using the definition of the pessimistic set, there exists $z_3 \in \mathcal{A}_{s_3}$ such that $\bs{R}_{s_3}(z_3) + C \subsetneq \bs{R}_{s_3}(z_2) + C$, which implies that
\begin{equation}\label{eq:143}
   \bs{R}_{s_3}(z_3)\subseteq \bs{R}_{s_3}(z_3)+ C \subseteq \bs{R}_{s_3}(z_2)+C \subseteq \bs{R}_{s_2}(z_2)+C~,
\end{equation}
where the last inclusion follows from the fact that $\bs{R}_{s_3}(z_2) \subseteq \bs{R}_{s_2}(z_2)$ as $s_3\geq s_2$. Then, combining \eqref{def:pesssetule} and \eqref{eq:143} and using the fact that $\bs{R}_{s_2}(z_1)\subseteq \bs{R}_{s_1}(z_1)$ yield
\[
\bs{R}_{s_3}(z_3)\subseteq \bs{R}_{s_2}(z_2)+C\subseteq \bs{R}_{s_2}(z_1)+C+C=\bs{R}_{s_2}(z_1)+C\subseteq \bs{R}_{s_1}(z_1)+C.
\]
Then, by \eqref{eq:hhj}, we get
\begin{align*}
    \bs{R}_{s_3}(z_3) + \epsilon\bs{u^*} \subseteq \bs{R}_{s_1}(z_1)+\epsilon\bs{u^*}+C
    \subseteq \bs{f}(x) + C ~.
\end{align*}
Since $\bs{f}(z_3) \in \bs{R}_{s_3}(z_3)$ under event $E$, we have
\begin{align*}
\bs{f}(z_3) + \epsilon \bs{u^*} &\in \bs{f}(x) + C ~.
\end{align*}
So, if $z_3 \in \hat{\mathcal{P}}$, then the claim holds with $ z = z_3$.

\rev{If $z_3 \not\in \hat{\mathcal{P}}$, then a similar argument can be made until $z_n \in \hat{\mathcal{P}}$, where $n\leq \abs{\mathcal{X}}$. Indeed, in the worst case, there exists some $z_n \in \mathcal{P}_{\text{pess},t_s}$ and it must be the case that $z_n 
 \in \hat{\mathcal{P}}$ since the algorithm terminates at round $t_s$. Then, the claim holds with $ z = z_n$. \qed}


\paragraph{} \rev{
The next lemma shows that if a design $x$ has a gap $\Delta_x^*>2 \epsilon$, then it cannot be in the predicted Pareto set $\hat{\mathcal{P}}$. To show this by contradiction, suppose such an $x$ is included in $\hat{\mathcal{P}}$. By the definition of $\Delta_{x'}^*$, there must exist a Pareto optimal design $x' \in P^*$ such that $m\left(x, x'\right)>2 \epsilon$.
If $x'$ is active when $x$ is added to $\hat{\mathcal{P}}$, Pareto identification rule (Equation \ref{eq:epseq}) leads to a contradiction. If $x'$ is not active, we consider two further cases: (i) if the design that discarded $x'$ was active when $x$ was added to $\hat{\mathcal{P}}$, the Pareto identification rule leads to a contradiction again; (ii) if the mentioned design was not active, it must have been discarded and thus removed from the pessimistic Pareto set by another design. In this case, this process continues iteratively, until the final design in this sequence is active at the round at which $x$ is added to $\hat{\mathcal{P}}$, again leading to a contradiction via the Pareto identification rule.
}

\begin{lem}\label{lem:returnspareto2}
    Under event $E$, the set $\hat{\mathcal{P}}$ returned by VOGP satisfies condition (ii) in Definition~\ref{defn:success}.
\end{lem}

\textit{Proof.} We will show that if $\Delta_x^* > 2\epsilon$, then $x \notin \hat{\mathcal{P}} \setminus P^*$. To prove this by contradiction, suppose that $\Delta_x^* > 2\epsilon$ for a design $x \in \hat{\mathcal{P}} \setminus P^*$. By definition of $m(x,x')$, this means that there exists $x' \in P^*$ such that $m(x, x') > 2\epsilon$. Since $x \in \hat{\mathcal{P}}$, it must have been added to $\hat{\mathcal{P}}$ at some round $t$. By the $\epsilon$-covering rule of VOGP, that means for all $z \in \mathcal{W}_t$, $(\bs{R}_{t}(x)+ \epsilon \bs{u}^*+C) \cap(\bs{R}_{t}(z)-C) = \emptyset$. \\

We complete the proof by considering two cases:
\begin{itemize}
    \item \textbf{Case 1}: $x' \in \mathcal{W}_t$.
    \item \textbf{Case 2}: $x' \notin \mathcal{W}_t$.
\end{itemize}

\textbf{Case 1}: If $x' \in \mathcal{W}_t$, then
\begin{align}\label{eq:epsiloncoveringjhat_supp}
     (\bs{R}_{t}(x)+\epsilon\bs{u}^*+C) \cap (\bs{R}_{t}(x')-C) = \emptyset ~.
\end{align}
By the properties of the Minkowski sum, we have $\bs{R}_{t}(x)+\epsilon\bs{u}^* \subseteq \bs{R}_{t}(x)+\epsilon\bs{u}^*+C$. This together with (\ref{eq:epsiloncoveringjhat_supp}) results in
\begin{align}\label{eq:epsiloncoveringjhatsubset_supp}
     (\bs{R}_{t}(x)+\epsilon\bs{u}^*) \cap(\bs{R}_{t}(x')-C) = \emptyset ~.
\end{align}

According to Lemma~\ref{lem:prob}, under the good event $E$, $\bs{f}(x) \in \bs{R}_t(x)$ and $\bs{f}(x') \in \bs{R}_t(x')$. Combining this with (\ref{eq:epsiloncoveringjhatsubset_supp}), we conclude that
\begin{align}\label{eq:epsiloncoveringjhatsubset2_supp}
     \bs{f}(x)+\epsilon\bs{u^*} \notin \bs{f}(x')-C ~.
\end{align}
Because $\bs{f}(x)+ \epsilon\bs{u}^* \notin \bs{f}(x')- \text{int}(C)$, by the definition of $m(x, x')$, $m(x, x') \leq \epsilon$ and we get a contradiction for the case of $x' \in \mathcal{W}_t$. \\

\textbf{Case 2}:
If $x' \notin \mathcal{W}_t$, it must have been discarded at an earlier round $s_1 < t$. By the discarding rule, $\exists z_1 \in \mathcal{P}_{\text{pess},s_1}$ such that
\begin{align}\label{def:discardingrule_supp}
    & \bs{R}_{s_1}(z_1)+ \epsilon\bs{u}^* \subseteq \bs{y} + C, ~\forall \bs{y} \in \bs{R}_{s_1}(x') ~.
\end{align}
We proceed in Case 2, by considering the following two cases based on the status of $z_1$:
\begin{itemize}
    \item \textbf{Case 2.1}: $z_1 \in \mathcal{W}_t$.
    \item \textbf{Case 2.2}: $z_1 \notin \mathcal{W}_t$.
\end{itemize}

\textbf{Case 2.1}: If $z_{1} \in \mathcal{W}_t$, since $\bs{R}_{t}(z_1) \subseteq \bs{R}_{s_1}(z_1)$, (\ref{def:discardingrule_supp}) also implies that
\begin{align}
	& \bs{R}_{t}(z_1) + \epsilon\bs{u}^* \subseteq \bs{y} + C, ~\forall \bs{y} \in \bs{R}_{s_1}(x') \notag \\ \iff & \forall \bs{y}_{x',s_1} \in \bs{R}_{s_1}(x'), ~\forall \bs{y}_{z_1,t} \in \bs{R}_{t}(z_1) :
	\bs{y}_{z_1,t} \in \bs{y}_{x',s_1} - \epsilon\bs{u}^* + C \notag \\
	\iff & \forall \bs{y}_{x',s_1} \in \bs{R}_{s_1}(x'), ~\forall \bs{y}_{z_1,t} \in \bs{R}_{t}(z_1) : 
	\bs{y}_{x',s_1} - \epsilon\bs{u}^* \preccurlyeq_C \bs{y}_{z_1,t} ~.  \label{eq:finito1_supp}
\end{align}
Since $z_{1} \in \mathcal{W}_t$, the $\epsilon$-covering rule at round $t$ between $x$ and $z_1$ pairs should hold. Combined with the fact that $\bs{R}_{t}(x)+ \epsilon \bs{u}^* \subseteq \bs{R}_{t}(x)+\epsilon\bs{u}^*+C$, it holds that
\begin{align}
     & (\bs{R}_{t}(x)+\epsilon\bs{u}^*+C) \cap (\bs{R}_{t}(z_1)-C) = \emptyset \\
     \implies&(\bs{R}_{t}(x)+\epsilon\bs{u}^*) \cap(\bs{R}_{t}(z_{1})-C) = \emptyset \notag \\
     \iff& \forall \bs{y}_{z_1,t} \in \bs{R}_{t}(z_1), \forall \bs{y}_{x,t} \in \bs{R}_{t}(x) : \bs{y}_{x,t}+ \epsilon \bs{u}^* \not\preccurlyeq_C \bs{y}_{z_1,t} ~. \label{eq:finito2_supp}
\end{align}
Then, by combining (\ref{eq:finito1_supp}) and (\ref{eq:finito2_supp}), we get $\bs{y}_{x,t}+ \epsilon \bs{u}^* \not\preccurlyeq_C \bs{y}_{x',s_1} - \epsilon\bs{u}^*$, $\forall \bs{y}_{x,t} \in \bs{R}_{t}(x)$ and $\forall \bs{y}_{x',s_1} \in \bs{R}_{s_1}(x')$. Therefore, according to Lemma \ref{lem:prob}, under the good event $\mathcal{F}_1$ it holds that $\bs{f}(x)+2 \epsilon  \bs{u}^* \not\preccurlyeq_C \bs{f}(x')$. Since $2\epsilon \bs{u}^* \in \mathbb{B}(2\epsilon) \cap C$, by the definition of $m(x, x')$, $m(x, x') \leq 2\epsilon$ which is a contradiction.

\textbf{Case 2.2}: 
Next, we examine the case $z_{1} \notin \mathcal{W}_t$. Particularly, consider the collection of designs denoted by $z_1, \dots, z_{n-1}, z_n$ where $z_i$ has been discarded at some round $s_{i+1}$ by being removed from $\mathcal{P}_{\text{pess},s_{i+1}}$ by $z_{i+1}$, as they fulfill the condition $\bs{R}_{s_{i+1}}(z_{i+1}) + C \subsetneq \bs{R}_{s_{i+1}}(z_{i}) + C$. Notice that this implies $\bs{R}_{s_{i+1}}(z_{i+1}) \subseteq \bs{R}_{s_{i+1}}(z_{i}) + C$. Assume that $z_n \in \mathcal{W}_{t}$. Notice that it's always possible to identify such a design, because of the way the sequence was defined. Together with \eqref{eq:inter}, we observe that the following set relations hold.
\begin{align}
     &\bs{R}_{s_{2}}(z_{1}) + C \supseteq \bs{R}_{s_{2}}(z_{2}) \supseteq \bs{R}_{s_{3}}(z_{2}) \notag \\
     &\bs{R}_{s_{3}}(z_{2}) + C \supseteq \bs{R}_{s_{3}}(z_{3}) \supseteq \bs{R}_{s_{4}}(z_{3}) \notag \\
     &\vdots \notag \\
     &\bs{R}_{s_{n-1}}(z_{n-2}) + C \supseteq \bs{R}_{s_{n-1}}(z_{n-1}) \supseteq \bs{R}_{s_{n}}(z_{n-1})\notag  \\
     &\bs{R}_{s_{n}}(z_{n-1})+C \supseteq \bs{R}_{s_{n}}(z_{n}) \supseteq \bs{R}_{t}(z_{n}) ~.\label{eq:pes2}
\end{align}
In particular, (\ref{eq:pes2}) holds since $t\geq s_n$. \\

By the definition of Minkowski sum and the convexity of $C$, for any $i \in \{2,\ldots,n-1\}$, we have 
\begin{align*}
    \bs{R}_{s_{i}}(z_{i-1}) + C \supseteq \bs{R}_{s_{i+1}}(z_{i})
    &\implies  \bs{R}_{s_{i}}(z_{i-1}) + C+C \supseteq \bs{R}_{s_{i+1}}(z_{i}) + C \\ 
    &\implies  \bs{R}_{s_{i}}(z_{i-1}) + C \supseteq \bs{R}_{s_{i+1}}(z_{i}) + C ~.
\end{align*}
Hence, it holds that
\begin{align}
     &\bs{R}_{s_{2}}(z_{1}) + C \supseteq \bs{R}_{s_{3}}(z_{2})  + C\notag \\
     &\bs{R}_{s_{3}}(z_{2}) + C \supseteq \bs{R}_{s_{4}}(z_{3})  + C\notag \\
     &\vdots \notag \\
     &\bs{R}_{s_{n-1}}(z_{n-2}) + C \supseteq \bs{R}_{s_{n}}(z_{n-1}) + C \notag \\
     \implies &\bs{R}_{s_{2}}(z_{1}) + C \supseteq \bs{R}_{s_{n}}(z_{n-1}) + C~. \label{eq:pes1}
\end{align}
Using the fact that $\bs{R}_{s_{1}}(z_{1}) \supseteq \bs{R}_{s_{2}}(z_{1})$ when $s_1\leq s_2$, (\ref{eq:pes1}), and (\ref{eq:pes2}) in order, we obtain
\begin{align}
    \bs{R}_{s_{1}}(z_{1})+C \supseteq \bs{R}_{s_{2}}(z_{1})+C \supseteq \bs{R}_{s_{n}}(z_{n-1}) + C \supseteq \bs{R}_{t}(z_{n}) ~. \label{eq:ntoone}
\end{align}

Next, by combining (\ref{def:discardingrule_supp}) and (\ref{eq:ntoone}), and using the properties of Minkowski sum, we have 
\begin{align}
    &\forall \bs{y} \in \bs{R}_{s_1}(x'): \bs{R}_{s_1}(z_1)+\epsilon \bs{u}^* \subseteq \bs{y} + C  
    \iff 
     \forall \bs{y} \in \bs{R}_{s_1}(x'): \bs{R}_{s_1}(z_1) \subseteq \bs{y}-\epsilon \bs{u}^* + C  \notag \\
     &\implies \forall \bs{y} \in \bs{R}_{s_1}(x'): \bs{R}_{s_1}(z_1)+C \subseteq \bs{y}-\epsilon \bs{u}^* + C \notag \\
    & \implies \forall \bs{y} \in \bs{R}_{s_1}(x'): \bs{R}_{t}(z_{n}) \subseteq  \bs{y}-\epsilon \bs{u}^* + C ~. \label{eq:frompes}
\end{align}
Alternatively, (\ref{eq:frompes}) can be re-written as
\begin{align}
 &\forall \bs{y}_{x',s_1} \in \bs{R}_{s_1}(x'), \forall \bs{y}_{z_n,t} \in \bs{R}_{t}(z_n):  \label{eq:finito1_suppfinal} \bs{y}_{x',s_1}-\epsilon \bs{u}^* \preccurlyeq_C  \bs{y}_{z_n,t} ~.
\end{align}
Since $z_{n} \in \mathcal{W}_t$, the $\epsilon$-covering rule at round $t$ between $x$ and $z_n$ pairs should hold. Combined with  $\bs{R}_{t}(x)+\epsilon \bs{u}^* \subseteq  \bs{R}_{t}(x)+\epsilon \bs{u}^*+C$, it holds that
\begin{align}
    & (\bs{R}_{t}(x)+\epsilon \bs{u^*}+C) \cap(\bs{R}_{t}(z_{n})-C) = \emptyset \\
     \implies &(\bs{R}_{t}(x)+\epsilon \bs{u}^*) \cap(\bs{R}_{t}(z_{n})-C) = \emptyset \notag \\
     \implies & \forall \bs{y}_{z_n,t} \in \bs{R}_{t}(z_n), \forall \bs{y}_{x,t} \in \bs{R}_{t}(x): \bs{y}_{x,t}+\epsilon \bs{u}^*  \not\preccurlyeq_C \bs{y}_{z_n,t}~.\label{eq:finito2_suppfinal}
\end{align}
Then, by combining (\ref{eq:finito1_suppfinal}) and (\ref{eq:finito2_suppfinal}), we get $\bs{y}_{x,t}+\epsilon \bs{u}^* \not\preccurlyeq_C \bs{y}_{x',s_1}-\epsilon \bs{u}^*$, $\forall \bs{y}_{x,t} \in \bs{R}_{t}(x)$ and $\forall \bs{y}_{x',s_1} \in \bs{R}_{s_1}(x')$.  Then, according to Lemma \ref{lem:prob}, under the good event $E$ it holds that $\bs{f}(x)+2\epsilon \bs{u}^* \not\preccurlyeq_C \bs{f}(x')$. Since $2\epsilon \bs{u}^* \in \mathbb{B}(2\epsilon) \cap C$, by the definition of $m(x, x')$, $m(x, x') \leq 2\epsilon$ which is a contradiction.\qed 

\rev{The following two lemmas show that, as confidence regions shrink over rounds, a point is reached where domination relations between any two points within a confidence hyperrectangle hold when $\epsilon \bs{u}^*$ help is included. This ensures that the algorithm can make all the necessary decisions, leading to eventual termination. Specifically, we prove that given that all the confidence hyperrectangles have shrunk below a cone dependent threshold, at that round, any design that is not to be added in the predicted Pareto set must have been discarded. Consequently, the algorithm terminates.}

\begin{lem}\label{lem:ustar}
     Let $t\in\mathbb{N}$ and suppose that $\overline{\omega}_t = \max_{x \in {\cal W}_t} \omega_t(x) < \frac{\epsilon}{d_C}$. Then, for every $x \in {\cal W}_t$ and $\bs{y},\bs{y}'\in \bs{R}_t(x)$, it holds
     \[
     \bs{y} \prec_C \bs{y}' + \epsilon \bs{u}^*.
     \]
\end{lem}

\textit{Proof.} Let $x\in{\cal W}_t$ and $\bs{y},\bs{y}'\in \bs{R}_t(x)$. By the definition of $\omega_t(x)$, we have
\[
    \bs{\tilde{p}}:=\bs{y}'-\bs{y} \in  \Int\left(\mathbb{B}\left(\frac{\epsilon}{d_C}\right)\right) ~.
\]
Moreover, we have $\bs{y}+\bs{\tilde{p}}=\bs{y}'\preccurlyeq_C\bs{y}'$ trivially. Hence, by Lemma~\ref{lem:ter}(ii), we directly get $\bs{y}\prec_C\bs{y}'+\epsilon\bs{u}^*$.
\qed 

\begin{lem}\label{lem:terminationcondition_withdirection}
   Let $t\in\mathbb{N}$ and suppose that $\overline{\omega}_t < \frac{\epsilon}{d_C(1)}$. Then, VOGP terminates at round $t$.
\end{lem}

\textit{Proof.} Let $\mathcal{S}_{t, 0}$  $(\mathcal{P}_{t, 0})$, $\mathcal{S}_{t, 1}$ $(\mathcal{P}_{t, 1})$, and $\mathcal{S}_{t, 2}$  $(\mathcal{P}_{t, 2})$ denote the set $\mathcal{S}_t$ $(\mathcal{P}_t)$ at the end of modeling, discarding and Pareto identification phases, respectively. 
For a node $x \in {\cal S}_{t,0} \setminus \mathcal{P}_{t,2}$, if $x \not\in \mathcal{S}_{t,1}$, then it must have been discarded at round $t$. Hence, VOGP terminates at round $t$ if and only if every $x \in {\cal S}_{t,0}$ is either discarded or moved to $\mathcal{P}_{t,2}$. In order to prove this, we show that if $x \in {\cal S}_{t,0} \setminus \mathcal{P}_{t,2}$ holds, then $x\notin\mathcal{S}_{t,1}$. To prove this by contradiction, let $x\in {\cal S}_{t,0}\setminus {\cal P}_{t,2}$ and assume that $x \in \mathcal{S}_{t,1}$. Since $x$ has not been added to $\mathcal{P}_{t,2}$, according to the $\epsilon$-covering rule in Algorithm~\ref{alg:pareto}, there exists some $ z^* \in \mathcal{P}_{t,1} \cup {\cal S}_{t,1}$ such that $(\bs{R}_t(x)+\epsilon\bs{u}^*+C)\cap \bs{R}_t(z^*)\neq\emptyset$. Hence, there exist $\bs{y}_{z^*} \in \bs{R}_{t}(z^*)$ and $\bs{y}_x \in \bs{R}_{t}(x)$ such that
\begin{align}
    \label{eq:ecovnew}
   \bs{y}_x + \epsilon\bs{u}^* \preccurlyeq_C \bs{y}_{z^*}~.
\end{align}
Since we assume that $x \in \mathcal{S}_{t,1} \subseteq \mathcal{P}_{t,1} \cup {\cal S}_{t,1} = {\cal W}_t$, by Lemma~\ref{lem:ustar}, we have
\begin{align}
    \forall \tilde{\bs{y}}_{x} \in \bs{R}_{t}(x): \tilde{\bs{y}}_{x} \prec_C \bs{y}_x+\epsilon \bs{u}^* ~. \label{eq:col1}
\end{align}
Again, by Lemma~\ref{lem:ustar}, since $z^* \in {\cal W}_t$, we have
\begin{equation}
    \forall \Tilde{\bs{y}}_{z^*} \in \bs{R}_{t}(z^*): \bs{y}_{z^*} \prec_C \Tilde{\bs{y}}_{z^*} +\epsilon \bs{u}^* ~.\label{eq:col2}
\end{equation}
Then, using \eqref{eq:ecovnew} using \eqref{eq:col2}, we have
\begin{equation}\label{eq:uy1}
    \forall \Tilde{\bs{y}}_{z^*} \in \bs{R}_{t}(z^*)\colon \bs{y}_x + \epsilon\bs{u}^* \preccurlyeq_C \bs{y}_{z^*} 
    \prec_C \Tilde{\bs{y}}_{z^*}  +\epsilon\bs{u}^*,
    \end{equation}
which implies that
\[
\forall \Tilde{\bs{y}}_{z^*} \in \bs{R}_{t}(z^*)\colon \bs{y}_x \prec_C \Tilde{\bs{y}}_{z^*} 
\]
so that
\begin{equation}
\bs{R}_t(z^*)\subseteq \bs{y}_x+\Int(C)~, \notag
\end{equation}
which then implies that
\[
\bs{R}_t(z^*)+C\subseteq \bs{y}_x+\Int(C)+C=\bs{y}_x+\Int(C).
\]
In particular, we have $\bs{R}_t(z^*)+C\subsetneq \bs{y}_x+C$. 
Together with $\bs{y}_x+C \subseteq \bs{R}_t(x)+C$, we get
\begin{equation}\label{eq:uy2}
\bs{R}_t(z^*)+C \subsetneq \bs{R}_t(x)+C ~.    
\end{equation}

On the other hand, using \eqref{eq:ecovnew} and \eqref{eq:col1}, we have 
\[
    \forall \tilde{\bs{y}}_{x} \in \bs{R}_{t}(x)\colon \tilde{\bs{y}}_{x} \prec_C \bs{y}_x+\epsilon \bs{u}^* \preccurlyeq_C \bs{y}_{z^*},
\]
which implies that
\begin{equation}\label{eq:neww}
   \forall \tilde{\bs{y}}_{x} \in \bs{R}_{t}(x)\colon  \tilde{\bs{y}}_{x} \prec_C \bs{y}_{z^*}~.
\end{equation}
Then, by \eqref{eq:neww} and \eqref{eq:col2}, we obtain
\[
    \forall \tilde{\bs{y}}_{x} \in \bs{R}_{t}(x)\ \forall \Tilde{\bs{y}}_{z^*} \in \bs{R}_{t}(z^*)\colon \tilde{\bs{y}}_{x} \prec_C \bs{y}_{z^*} \prec_C \Tilde{\bs{y}}_{z^*} +\epsilon \bs{u}^*~,
\]
which implies that
\[
 \forall \tilde{\bs{y}}_{x} \in \bs{R}_{t}(x)\ \forall \Tilde{\bs{y}}_{z^*} \in \bs{R}_{t}(z^*)\colon \tilde{\bs{y}}_{x} \preccurlyeq_C \Tilde{\bs{y}}_{z^*} +\epsilon \bs{u}^*
\]
so that
\begin{equation}\label{eq:row3}
 \forall \tilde{\bs{y}}_{x} \in \bs{R}_{t}(x)\colon \bs{R}_t(z^*)+\epsilon \bs{u}^*\subseteq \tilde{\bs{y}}_{x}+C~.
\end{equation}

 Then, by \eqref{eq:uy2}, we obtain $x\notin \mathcal{P}_{\text{pess},t}$. If $z^* \in \mathcal{P}_{\text{pess},t}$, then  \eqref{eq:row3} shows that $x$ must be discarded. If $z^* \notin \mathcal{P}_{\text{pess},t}$, then  $\exists z^\prime \in \mathcal{A}_t$ such that 
\begin{align}
    &\forall \bs{y}_{z^\prime} \in \bs{R}_t(z^\prime), \exists \bs{y}_{z^{\prime\prime}} \in \bs{R}_t(z^*): \bs{y}_{z^{\prime\prime}} \preccurlyeq_C \bs{y}_{z^\prime}  \label{eq:pess132} \\
    & \iff \forall \bs{y}_{z^\prime} \in \bs{R}_t(z^\prime), \exists \bs{y}_{z^{\prime\prime}} \in \bs{R}_t(z^*): \bs{y}_{z^{\prime\prime}}+\epsilon \bs{u}^* \preccurlyeq_C \bs{y}_{z^\prime}  +\epsilon \bs{u}^*\label{eq:pesslast} ~.
\end{align}
Fix $\bs{y}_{z^{\prime\prime}}$ as given in (\ref{eq:pess132}). Starting from (\ref{eq:row3}) and using (\ref{eq:pesslast}), we have
\begin{align}
     &\bs{y}_{\Tilde{x}} \preccurlyeq_C \bs{y}_{z^{\prime\prime}}+\epsilon \bs{u}^* \preccurlyeq_C \bs{y}_{z^\prime}  +\epsilon \bs{u}^*, \forall \bs{y}_{z^\prime} \in \bs{R}_t(z^\prime) \text{ and } \forall \bs{y}_{\Tilde{x}} \in \bs{R}_{t}(x)\\
     &\implies \bs{y}_{\Tilde{x}} \preccurlyeq_C \bs{y}_{z^\prime}  +\epsilon \bs{u}^*,\forall \bs{y}_{z^\prime} \in \bs{R}_t(z^\prime) \text{ and } \forall \bs{y}_{\Tilde{x}} \in \bs{R}_{t}(x) ~.\label{eq:tre32}
\end{align}
\eqref{eq:tre32} shows that if $z^\prime \in \mathcal{P}_{\text{pess},t}$, $x$ should be discarded. If $z^\prime \notin \mathcal{P}_{\text{pess},t}$, a similar argument can be made until the condition to be inside $\mathcal{P}_{\text{pess},t}$ holds since $\mathcal{A}_{t}$ is a finite set and $\mathcal{P}_{\text{pess},t}$ is not empty. Hence, the lemma is proved.\qed

\rev{The following lemma establishes an upper bound on the sum of maximum diagonal distances of the confidence regions over rounds, linking it to the maximum information gain. The proof follows by relating the maximum diagonal distance of confidence regions to the sum of variances and leveraging known bounds on information gain.  Finally, by applying the stopping condition from Lemma~\ref{lem:terminationcondition_withdirection}, we obtain a bound on the number of rounds required for termination, completing the proof of Theorem~\ref{thm:main}.}

\begin{lem}\label{lem:infogap} Let  $t_s$ represent the round in which the algorithm terminates. We have
	\begin{align*}
	\sum_{t=1}^{t_s} \overline{\omega}_{t} \leq \sqrt{t_s \left( 8\beta_{{t_s}}  \sigma^2 \eta M \gamma_{t_s} \right) }~,
	\end{align*}
	where $\eta =\frac{\sigma^{-2}}{\ln (1+\sigma^{-2})}$ and $\gamma_{t_s}$ is the maximum information gain in $t_s$ evaluations. \
\end{lem} 

\textit{Proof.} Since the diagonal distance of the hyperrectangle $\bs{Q}_{t}(x)$ is the largest distance between any two points in the hyperrectangle, we have
\begin{align}
  \sum_{t=1}^{t_s} \overline{\omega}^2_{t} 
&=\sum_{t=1}^{t_s}\max_{\bs{y},\bs{y}^\prime\in \bs{R}_{t}(x_t)}\norm{\bs{y}-\bs{y}^\prime}_2^2 \label{eq:3453}\\
&\leq \sum_{t=1}^{t_s}\max_{\bs{y},\bs{y}^\prime\in \bs{R}_{t-1}(x_t)}\norm{\bs{y}-\bs{y}^\prime}_2^2  \notag\\
&\leq  \sum_{t=1}^{t_s}\max_{\bs{y},\bs{y}^\prime\in \bs{Q}_{t-1}(x_t)}\norm{\bs{y}-\bs{y}^\prime}_2^2\notag\\
& =   \sum_{t=1}^{t_s} \sum^M_{j=1} \left( 2 \beta^{1/2}_{t-1} \sigma^j_{t-1}(x_t ) \right)^2 \label{eq:hyperr}\\
& \leq 4\beta_{t_s}  \sum_{t=1}^{t_s} \sum^M_{j=1}   (\sigma^j_{t-1}(x_t))^2 \label{eq448}\\
& = 4\beta_{{t_s}}  \sigma^2   \sum_{t=1}^{t_s} \sum^M_{j=1}   \sigma^{-2} (\sigma^j_{t-1}(x_t))^2 \notag \\
& \leq 4\beta_{{t_s}}  \sigma^2 \eta \left(  \sum_{t=1}^{t_s} \sum^M_{j=1} \ln (1 +  \sigma^{-2}  (\sigma^j_{t-1}(x_t))^2) \right) \label{eq450}\\
& \leq 8\beta_{{t_s}}  \sigma^2 \eta M I( \bs{y}_{[{t_s}]}; f_{[{t_s}]}) \label{eq451}\\
& \leq 8\beta_{{t_s}}  \sigma^2 \eta M \gamma_{t_s},\label{eq452}
\end{align}
where $\eta := \sigma^{-2} / \ln \left(1+\sigma^{-2}\right)$ and $\sigma$ is the noise standard deviation; (\ref{eq:3453}) is due to the definition of $\bar{\omega}_t$; (\ref{eq:hyperr}) follows from (\ref{eq:q}); (\ref{eq448}) holds since $\beta_t$ is nondecreasing in $t$; (\ref{eq450}) follows from the fact that $s \leq \eta \ln (1+s)$ for all $0 \leq s \leq \sigma^{-2}$ and that we have
\begin{align}
& \sigma^{-2}\left(\sigma_{t-1}^j\left(x_t\right)\right)^2  \notag \\
& = \sigma^{-2} \Bigl( k^{j j}\left(x_t, x_t\right)-\left(\bs{k}_{[t-1]}\left(x_t\right)\left(\bs{K}_{[t-1]}+\bs{\Sigma}_{[t-1]}\right)^{-1} \bs{k}_{[t-1]}\left(x_t\right)^{\mathsf{T}}\right)^{j j} \Bigr) \notag\\
&\leq \sigma^{-2} k^{j j}\left(x_t, x_t\right)\notag \\
&\leq \sigma^{-2} ,\label{eq:boundedvar}
\end{align}
where \eqref{eq:boundedvar} follows from the fact that $k^{jj}(x,x') \leq 1$ for all $x,x' \in {\cal X}$, and (\ref{eq451}) follows from \citet[Proposition~1]{nika2021pareto}. Finally, by Cauchy-Schwarz inequality, we have
\begin{align*}
\sum_{t=1}^{t_s} \overline{\omega}_{t} \leq \sqrt{ {t_s}  \sum_{t=1}^{t_s} \overline{\omega}^2_{t}}\leq \sqrt{{t_s} \left(  8\beta_{{t_s}}  \sigma^2 \eta M \gamma_{t_s} \right) } ~. 
\end{align*}\qed

By definition, $\overline{\omega}_t = \omega_t(x_t) \leq \omega_{t-1}(x_t) \leq \max_{x \in {\cal W}_{t-1}} \omega_{t-1}(x) = \overline{\omega}_{t-1}$, where the first inequality is due to $\bs{R}_t(x)\subseteq \bs{R}_{t-1}(x)$ for $x \in {\cal X}$ and the last inequality is due to the selection rule of VOGP. Hence, we conclude that $\overline{\omega}_t \leq \overline{\omega}_{t-1}$. By above and by Lemma~\ref{lem:infogap}, we have
	\begin{align*}
	\overline{\omega}_{t_s} \leq \frac{\sum^{t_s}_{t=1} \overline{\omega}_{t} }{t_s} \leq  \sqrt{\frac{8\beta_{t_s}  \sigma^2 \eta M \gamma_{t_s} }{t_s}} ~.
	\end{align*}

By using the stopping condition of Lemma \ref{lem:terminationcondition_withdirection}, we get 

\begin{equation}
     T \coloneqq \min \left\{ t \in \mathbb{N} :  \sqrt{\frac{8\beta_{t}  \sigma^2 \eta M \gamma_{t} }{t}} < \frac{\epsilon}{d_C(1)} \right\} ~. \nonumber 
\end{equation}
Hence, the proof of Theorem \ref{thm:main} is complete.

\subsection{Proof of Theorem \ref{thm:kernel}} \label{sec:thm2proof} 

The following auxiliary lemma will be helpful. 

\begin{lem}[\cite{Broxson_2006}, Theorem 15] \label{lem:kronocker} 
 Let $\bs{A} \in \mathbb{R}^{n\times n}$, $\bs{B} \in \mathbb{R}^{m\times m}$ be two real square matrices, where $m,n\in\mathbb{N}$. If $\lambda$ is an eigenvalue of $\bs{A}$ with corresponding eigenvector $\bs{v} \in \mathbb{R}^n$ and $\mu$ is an eigenvalue of $\bs{B}$ with corresponding eigenvector $\bs{u} \in \mathbb{R}^m$, then $\lambda \mu$ is an eigenvalue of $\bs{A} \otimes \bs{B}\in\mathbb{R}^{mn\times mn}$, the Kronecker product of $\bs{A}$ and $\bs{B}$, with corresponding eigenvector $\bs{v} \otimes \bs{u} \in \mathbb{R}^{mn}$. Moreover, the set of eigenvalues of $\bs{A} \otimes \bs{B}$ is $\{\lambda_i \mu_j: i\in[n], j\in[m]\}$, where $\lambda_1,\ldots,\lambda_n$ are the eigenvalues of $\bs{A}$ and $\mu_1,\ldots,\mu_m$ are the eigenvalues of $\bs{B}$ (including algebraic multiplicities). In particular, the set of eigenvalues of $\bs{A} \otimes \bs{B}$ is the same as the set of eigenvalues of $\bs{B} \otimes \bs{A}$. 
\end{lem}

The following lemma associates the information gain of an $M$-output GP with the maximum information gain of single output GPs.

\begin{lem}\label{lem:siribound}
    Let $\bs{f}$ be a realization from an $M$-output GP with a separable covariance function of the form $(x,x^\prime)\mapsto \bs{k}(x, x^{\prime})=[\tilde{k}(x, x^{\prime})k^*(p, q)]_{p, q \in[M]}$, where $\tilde{k}\colon\mathcal{X}\times\mathcal{X}\to\mathbb{R}$ is an RBF or Matérn kernel for the design space and $k^*\colon[M]\times[M]\to\mathbb{R}$ is a kernel for the objective space. 
    For each $p\in[M]$, let $\Psi_p$ be the maximum information gain for a single output GP whose kernel is $(x,x^\prime)\mapsto \tilde{k}(x,x^\prime)k^*(p,p)$. Then, for every $t\in\mathbb{N}$, we have
    \[
         I(\bs{y}_{[t]};\bs{f}_{[t]}) \leq  M  \max_{p \in [M]}\Psi_p ~.
    \]
\end{lem}

 \textit{Proof.}
Our proof is similar to the proof of \citet[Theorem~2]{li2022safe}. The difference comes from the structures of the covariance matrices, where the order of Kronecker products are swapped in our case. \rev{In other words,} Kronecker product of input and output kernels to form the covariance matrix has swapped order.
 
Recall that $\bs{K}_{t} =(\bs{k}\left(x_i, x_j\right))_{i,j\in[t]}$. Hence, we have
\begin{align*}
     I(\bs{y}_{[t]}; \bs{f}_{[t]})&=H(\bs{y}_{[t]})-H(\bs{y}_{[t]} \mid \bs{f}_{[t]})\\
    & =\frac{1}{2} \ln \left|2 \pi e\left(\bs{K}_{t}+\sigma^2 \bs{I}_{M t}\right)\right|-\frac{1}{2} \ln \left|2 \pi e \sigma^2 \bs{I}_{M t}\right| \\
    & =\frac{1}{2} \ln \left(\frac{\left|2 \pi e\left(\bs{K}_{t}+\sigma^2 \bs{I}_{M t}\right)\right|}{\left|2 \pi e \sigma^2 \bs{I}_{M t}\right|}\right)\\
    & =\frac{1}{2} \ln \left|\left(\bs{K}_{t}+\sigma^2 \bs{I}_{Mt}\right) \cdot\left(\sigma^2 \bs{I}_{Mt}\right)^{-1}\right| \\ 
    & =\frac{1}{2} \ln \left|\bs{I}_{Mt}+\frac{1}{\sigma^2} \bs{K}_{t}\right|.
\end{align*}
By the separable form of $\bs{k}$, we have
\[
\bs{K}_{t}=[\tilde{k}(x_i, x_j)]_{i, j \in[t]}\otimes\left[k^*(p, q)\right]_{p, q \in[M]}.
\]
Hence, by Lemma~\ref{lem:kronocker} and using the identity $\abs{ \bs{I} + \bs{A}\otimes\bs{B}} = \abs{ \bs{I} + \bs{B}\otimes\bs{A}}$, we get
\begin{align}
    I(\bs{y}_{[t]}; \bs{f}_{[t]})=\frac{1}{2} \ln \abs{ \bs{I}_{Mt}+\frac{1}{\sigma^2} 
     [k^*(p, q)]_{p, q \in[M]}\otimes [\tilde{k}(x_i, x_j)]_{i, j [t]} } ~. \nonumber
\end{align}
Notice that 
\begin{align*}
    &\bs{I}_{Mt}+\frac{1}{\sigma^2} 
     [k^*(q, p)]_{p, q \in[M]}\otimes [\tilde{k}(x_i, x_j)]_{i, j \in[t]}  \notag \\
    &=\left[\begin{array}{ccc}
\bs{I}_{t}+k^*(1,1)[\tilde{k}(x_i, x_j)]_{i, j \in[t]}\sigma^{-2}, & \ldots, & k^*(1,M)[\tilde{k}(x_i, x_j)]_{i, j \in[t]}\sigma^{-2} \\
\vdots & & \vdots \\
k^*(M,1)[\tilde{k}(x_i, x_j)]_{i, j \in[t]}\sigma^{-2}, & \ldots, & \bs{I}_{t}+k^*(M,M)[\tilde{k}(x_i, x_j)]_{i, j \in[t]}\sigma^{-2} 
\end{array}\right] ~.
\end{align*}
Since the matrix itself and all of its diagonal blocks are positive definite symmetric real matrices, we can apply Fischer's inequality and obtain
\begin{align*}
     I(\bs{y}_{[t]}; \bs{f}_{[t]}) \leq \frac{1}{2} \sum_{p=1}^M \ln \left|\bs{I}_{t}+k^*(p,p)\sigma^{-2}[\tilde{k}(x_i, x_j)]_{i, j \in [t]}\right|. \\
\end{align*}
This is actually the sum of mutual informations of single output GPs $f_l \sim \mathcal{G P} (\bs{0},k^*(l,l)[\tilde{k}(x_i, x_j)]_{i, j \in [t]})$. Notice that a positive constant multiple of an RBF (resp. Matérn) kernel is still an RBF (resp. Matérn) kernel.  Since the mutual information is bounded by maximum information gain, we obtain $I(\bs{y}_{[t]}; \bs{f}_{[t]}) \leq  M  \max_{p \in [M]}\Psi_p$, which completes the proof.\qed

\rev{Now, we can continue with the proof of Theorem \ref{thm:kernel}}. By Lemma \ref{lem:terminationcondition_withdirection}, we have to show that taking 
\[
    t= \frac{32 d_C^2}{\epsilon^2}  \phi M^2 \sigma^2 \eta \cdot 
    \ln^{D+2}\left(
        \sqrt{
            \frac{M|\mathcal{X}|}{3 \delta}
        } \pi \cdot \frac{32 d_C^2}{\epsilon^2}  \phi M^2 \sigma^2 \eta
    \right)
\]
satisfies $ \frac{8\beta_{t}  \sigma^2 \eta M \gamma_{t} }{t}  < \frac{\epsilon^2}{d_C^2(1)}$ where $\phi$ is the multiplicative constant that comes from $\mathcal{O}(\cdot)$ notation of the bound on single output GP's information gain. We will first upper-bound the LHS of this inequality and show that its smaller than or equal to the RHS of the inequality. We are using an RBF kernel $\tilde{k}$ for the design space. Then, by Lemma~\ref{lem:siribound} and the bounds on maximum information gain established in \cite{vakilibounds}, we have 

\[
    I(\boldsymbol{y}_{[t]}; \boldsymbol{f}_{[t]}) \leq  M \cdot \mathcal{O}\left(\ln^{D+1} (t)\right) 
     \implies \gamma_t \leq  M \cdot \mathcal{O}\left(\ln^{D+1} (t)\right) ~. \\
\]
Notice that in Theorem \ref{thm:main}, as $\epsilon$ goes to 0, $T$ goes to infinity. Therefore, we can use the bounds on maximum information gain established in \cite{srinivas2012information}.  We have $\beta_t =\ln (\frac{M\pi^2 |\mathcal{X}| t^2}{3\delta })$.

\begin{align*}
    \frac{8 \beta_t \sigma^2 \eta M \gamma_t}{t} &= \frac{8 \ln (\frac{M\pi^2 |\mathcal{X}| t^2}{3\delta }) \sigma^2 \eta M \gamma_t}{t} \\
     &\leq \frac{8 \ln (\frac{M\pi^2 |\mathcal{X}| t^2}{3\delta }) \sigma^2 \eta M^2 \phi \ln^{D+1}(t)}{t} \\
    & =  \frac{16 \ln \left(\sqrt{\frac{M|\mathcal{X}|}{3 \delta}} \cdot t\pi\right) \sigma^2 \eta M^2 \phi\ln^{D+1}(t)}{t} \\
    & <  \frac{16 \ln^{D+2} \left(\sqrt{\frac{M|\mathcal{X}|}{3 \delta}} \cdot t\pi\right) \sigma^2 \eta M^2 \phi}{t}
\end{align*}
where $\phi$ is the multiplicative constant that comes from the $\mathcal{O}(\cdot)$ notation. The last inequality follows from $M\geq 1, \delta \in (0,1), \abs{\mathcal{X}} \geq 1$ and $\pi > \sqrt{3}$. Now, we can plug in \[
    t= \frac{32 d_C^2}{\epsilon^2} M^2  \sigma^2 \eta \phi\cdot 
    \ln^{D+2}\left(
        \sqrt{
            \frac{M|\mathcal{X}|}{3 \delta}
        } \pi \cdot \frac{32 d_C^2}{\epsilon^2}\phi M^2  \sigma^2 \eta
    \right) 
\]
to the last expression:

\begin{align*}
    &\frac{16 \ln ^{D+2}\left(\sqrt{\frac{M|\mathcal{X}|}{3 \delta}} \pi \frac{32 d_C^2}{\epsilon^2}\phi  M^2  \sigma^2 \eta \cdot \ln ^{D+2}\left(\sqrt{\frac{M|\mathcal{X}|}{3 \delta}} \pi \cdot \frac{32 d_C^2}{\epsilon^2} \phi  M^2  \sigma^2 \eta\right)\right)\phi  M^2  \sigma^2 \eta}{\frac{32 d_C^2}{\epsilon^2} \phi  M^2  \sigma^2 \eta \cdot \ln ^{D+2}\left(\sqrt{\frac{M|\mathcal{X}|}{3 \delta}} \pi \frac{32 d_C^2}{\epsilon^2}\phi  M^2  \sigma^2 \eta\right)} \\
    & = \frac{\ln ^{D+2}\left(\sqrt{\frac{M|\mathcal{X}|}{3 \delta}} \pi \frac{32 d_C^2}{\epsilon^2}\phi  M^2  \sigma^2 \eta \cdot \ln ^{D+2}\left(\sqrt{\frac{M|\mathcal{X}|}{3 \delta}} \pi \cdot \frac{32 d_C^2}{\epsilon^2}\phi  M^2  \sigma^2 \eta\right)\right) }{ 2 \ln ^{D+2}\left(\sqrt{\frac{M|\mathcal{X}|}{3 \delta}} \pi \frac{32 d_C^2}{\epsilon^2}\phi  M^2  \sigma^2 \eta\right)} \cdot \frac{\epsilon^2}{  d_C^2}\\
\end{align*}

To show that the last expression is smaller than or equal to $\frac{\epsilon^2}{  d_C^2}$, we need to show that 
\[
\ln^{D+2}\left(B \ln^{D+2}(B)\right) \leq 2 \ln^{D+2}(B)=\ln ^{D+2}\left(B^{\left({2}^{\frac{1}{D+2}}\right)}\right)
\]
where $B = \sqrt{\frac{M|\mathcal{X}|}{3 \delta}} \pi \frac{32 d_C^2}{\epsilon^2}\phi  M^2  \sigma^2 \eta$ for convenience. This holds because $\ln^{D+2} B \leq B^{\left({2}^{\frac{1}{D+2}}-1\right)}$ for sufficiently large  $B$. Hence, the proof is complete. \qed

\section{Numerical results}\label{sec:exp}

In this section, we analyze where VOGP stands when compared to the state-of-the-art.

\subsection{Datasets, cones and performance benchmarks}

\noindent\textbf{BC} (Branin-Currin, $D=2$, $M=2$, $\lvert\mathcal{X}\rvert=500$) Simultaneous optimization of two widely used single-output test benchmarks.

\noindent\textbf{LAC} (Lactose, $D=2$, $M=2$, $\lvert\mathcal{X}\rvert=250$) A chemical reaction for the isomerization of lactulose from lactose \citep{HASHEMI2010181, D1RE00549A}.

\noindent\textbf{VS} (Vehicle Safety, $D=5$, $M=3$, $\lvert\mathcal{X}\rvert=500$) Optimization of vehicle structures concerning safety, i.e., crashworthiness \citep{vehiclesafety}.

\noindent\textbf{SnAr} ($D=4$, $M=2$, $\lvert\mathcal{X}\rvert=2000$) Optimization of the nucleophilic aromatic substitution reaction (SnAr) involving 2,4-difluoronitrobenzene and pyrrolidine \citep{C6RE00109B}.

\noindent\textbf{BCC} (Branin-Currin Continuous, $D=2$, $M=2$) We use the same test benchmark functions as in BC dataset with continuous domain over $[0, 1]^2$.

\noindent\textbf{ZDT3} ($D=2$, $M=2$) A commonly used test benchmark function with continuous domain over $[0, 1]^2$ \citep{zitzler2000comparison}.

In our experiments, we use three cones which we call \textit{acute}, \textit{right}, and \textit{obtuse}. For $M=2$, they correspond to $C_{60^\circ}$, $C_{90^\circ}$, and $C_{120^\circ}$ in Definition~\ref{def:2dcone} below, respectively. Figure~\ref{fig:cones} illustrates these ordering cones. However, for VC dataset, which is a tri-objective problem ($M=3$), we define acute and obtuse cones with $\bs{W}$ matrices given as 
\[
\text{Acute cone: }
\bs{W}=\sqrt{21}
\begin{bmatrix}
1 & -2 & 4 \\
4 & 1 & -2 \\
-2 & 4 & 1
\end{bmatrix}
\hfill
\text{, ~ Obtuse cone: }
\bs{W}=\sqrt{3.72}
\begin{bmatrix}
1 & 0.4 & 1.6 \\
1.6 & 1 & 0.4 \\
0.4 & 1.6 & 1
\end{bmatrix}
\]
alongside the positive orthant cone (right cone) with identity matrix.

\begin{defn}\label{def:2dcone}
    Given $\theta\in (0,180)$, the cone $C_{\theta^\circ} \subseteq \mathbb{R}^2$ is defined as the convex cone whose two boundary rays make $\frac{\pm\theta}{2}$ degrees with the identity line (see Figure~\ref{fig:cones}). 
\end{defn}

\begin{remark}
    Cone $C_{90^\circ}$ in Definition \ref{def:2dcone} corresponds to multi-objective optimization with $M=2$.
\end{remark}

\begin{figure}[h!]
     \centering
     \includegraphics[width=0.3\textwidth]{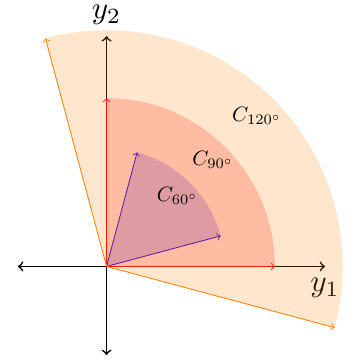}
     \caption{Acute, right and obtuse ordering cones for $M=2$.}
     \label{fig:cones}
 \end{figure}

Since our task is simply the classification of the Pareto designs, in the spirit of the widely used F1 score, we use the following definition that integrates our success condition (Definition~\ref{defn:success}) into a performance measure.

\begin{defn}
    \citep{karagozlu2024learning} Given a finite set of designs $\mathcal{X}$, set of Pareto designs $P^*$ and a set of predicted Pareto designs $P$ w.r.t. the ordering cone $C$, and an $\epsilon$, we define; a positive class as $\Pi_\epsilon = \{ x \in  \mathcal{X} : \Delta^*_x \le \epsilon \}$, where $\Delta^*_x$ is the suboptimality gap, and an $\epsilon$ lenient F1 classification score ($\epsilon$-F1) as
    \[
    \epsilon\text{-F1} = \frac{2 |\Pi_\epsilon \cap P|}{2 |\Pi_\epsilon \cap P| + |\Pi_{\epsilon \setminus P}| + |P \setminus \Pi_\epsilon|}~,
    \]
    where $\Pi_{\epsilon \setminus P}$ is the set of Pareto optimal arms that do not satisfy Definition \ref{defn:success} part (i).
\end{defn}

In multi-objective optimization, hypervolume indicator, denoted $\mathrm{HV}(\hat{\mathcal{P}})$, is an ubiquitous metric and it corresponds to the volume dominated by a set representing the found Pareto front, i.e., $\hat{\mathcal{P}}$, and a reference point. This metric is especially useful when working with continuous domains. Further, HV discrepancy, defined by
\[
d_{\mathrm{HV}} = \lvert \mathrm{HV}(\mathcal{P}^*) - \mathrm{HV}(\hat{\mathcal{P}}) \rvert,
\]
is a well-known measure of how well the set $\hat{\mathcal{P}}$ represents a known, true Pareto front, 
i.e., $\mathcal{P}^*$. Below, we extend the definition of HV to vector optimization. We also establish a theoretical equivalence with the original definition in the case where $M=N$, using the Jacobian of cone matrix $\boldsymbol{W}$.
\begin{defn} \label{def:conehyper}
    Given a reference point $\boldsymbol{r} \in \mathbb{R}^M$ and $N\times M$ matrix $\boldsymbol{W}$ of ordering cone $C$, hypervolume indicator $(\mathrm{HV}_C)$ of a finite approximate Pareto set $\hat{\mathcal{P}}$ is the $N$-dimensional Lebesgue measure $\lambda_N$ of the space dominated by $\mathcal{P}$ and bounded from below by $\boldsymbol{r}: \mathrm{HV}_C(\hat{\mathcal{P}}, \boldsymbol{r})=\lambda_N\left(\bigcup_{i=1}^{|\hat{\mathcal{P}}|}\left[\boldsymbol{W}\boldsymbol{r},\boldsymbol{W}\boldsymbol{y}_i\right]\right)$, where $\left[\boldsymbol{W}\boldsymbol{r},\boldsymbol{W}\boldsymbol{y}_i\right]$ denotes the hyperrectangle bounded by vertices $\boldsymbol{W}\boldsymbol{r}$ and $\boldsymbol{W}\boldsymbol{y}_i$.
\end{defn}

\begin{figure}[h!]
     \centering
     \includegraphics[width=1\textwidth]{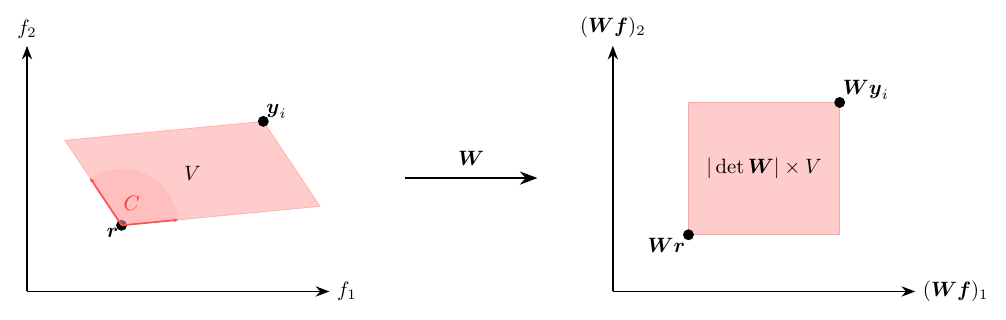}
     \caption{\revf{Illustration of Definition 7 in a simple two dimensional case. The left plot shows the dominated region $V$ under the cone $C$, and the right plot shows its image after the linear transformation induced by $\boldsymbol{W}$, where the region becomes axis-aligned and its volume scales by $|\operatorname{det} \boldsymbol{W}|$.}}
     \label{Fig:new2}
 \end{figure}

\begin{remark}
    The cone hypervolume $\mathrm{HV}_C$ defined in Definition \ref{def:conehyper} provides a general definition of hypervolume for any dimensions $N,M$. However, when $N=M$, by Theorem 2.3.15 of \citet{hypervolume}, the $\mathrm{HV}_C$ simplifies to $ |\det \boldsymbol{W}| \lambda_M\left(\bigcup_{i=1}^{|\hat{\mathcal{P}}|} 
 ( \boldsymbol{r} + C ) \cap (\bs{y}_i-C)  \right)$ where $\boldsymbol{r}\in \mathbb{R}^M$ is the reference point. This represents the hypervolume that is cone dominated by the Pareto front. \revf{A two dimensional illustration of this transformation and the corresponding hypervolume scaling is provided in Figure \ref{Fig:new2}.}
\end{remark}

\subsection{Experimental setup and competing algorithms}

We apply min-max normalization to inputs. We also standardize the outputs to work on a unified scale of $[0,1]$, i.e., for $\epsilon$ and $\sigma$. We fix $\epsilon=0.1$, $\delta=0.05$, $\sigma=0.1$. \revf{The GP model is initialized using a single randomly selected design point to start the optimization.} We use RBF kernel featuring automatic relevance determination for all experiments as it is fairly standard, and in line with our GP assumption in Section~\ref{sec:PD}, we assume known hyperparameters and choose them using maximum likelihood estimation prior to optimization. Similar to the other works with the same confidence hyperrectangle definition \citep{JMLR:v17:15-047}, we scale down $\beta_t$ by an empirical factor, in this case $2^5=32$. We report our results as average of 10 runs.

We use a series of algorithms to compare against. Previous work introduced Na\"ive Elimination (NE) \citep{ararat2023vector} algorithm that is designed for vector optimization using ordering cones. Even though it is an ($\epsilon,\delta$)-PAC algorithm, since its theoretical sampling budget is impractically high, we treat NE as a fixed-budget algorithm. It uses a per-design sampling budget $L$, and we set $L = \ceil{T/|\mathcal{X}|}$ where $T$ is the sampling complexity of VOGP for that specific experiment. Another work on vector optimization that does ($\epsilon,\delta$)-PAC Pareto set identification is PaVeBa \citep{karagozlu2024learning}. PaVeBa, samples all designs that are not discarded at all rounds. We scale down their confidence radius $r_t$ by 32 in this method to limit high sample complexities. As NE and PaVeBa also assume finite design set, we start by comparing VOGP with these two and report the results in Table~\ref{tab:vogp_vector_comp}. \rev{Also, we provide the convergence plots for the results in Table~\ref{tab:vogp_vector_comp} in Appendix \ref{supp:exp}.}
To introduce further baselines, we refer to multi-objective optimization literature, which is a special case of vector optimization where the ordering cone is the positive orthant cone. We devise fixed-budget variants for widely used information theoretic acquisition functions MESMO \citep{mesmo} and JES \citep{jesmo}.
As these algorithms work in continuous domains, we propose an adaptive discretization \citep{bubeck2011x} based VOGP as a variant to compare.
We run these acquisition functions over the design set $\mathcal{X}$ with GPs using the same kernel as VOGP and with the same sample complexity as VOGP. We compute the predicted Pareto set w.r.t. the ordering cone on the posterior mean for MESMO, JES and VOGP.
A naive approach of adapting these acquisition functions would be transforming the problem by applying the linear mapping induced by matrix $\boldsymbol{W}$ and then using the methods as if doing a multi-objective optimization. However, this approach proves ineffective and a more in depth discussion on this can be found in Section~\ref{app:w_transform}.
We do not compare VOGP with $\epsilon$-PAL since in the multi-objective setting, VOGP can be considered as an extension of it and adapting $\epsilon$-PAL to the ordering cones is non-trivial.

\begin{table*}[!t]
    \centering
    \resizebox{\columnwidth}{!}{
    \begin{tabular}{llcccccc}
        \toprule
        & & \multicolumn{2}{c}{Acute Cone} & \multicolumn{2}{c}{Right Cone} & \multicolumn{2}{c}{Obtuse Cone} \\
        \midrule
        Dataset & Alg. & S.C. & $\epsilon$-F1 ($\uparrow$) & S.C. & $\epsilon$-F1 ($\uparrow$) & S.C. & $\epsilon$-F1 ($\uparrow$) \\
        \midrule
        \multirow[t]{3}{*}{BC}
        & NE & 500 & 0.89 ± .07 & 500 & 0.92 ± .07 & 500 & 0.98 ± .06 \\
        & PaVeBa & 517.7 ± 5.3 & 0.95 ± .03 & 504.5 ± 3.0 & 0.96 ± .04 & 500.9 ± 1.1 & 0.93 ± .09 \\
        \cmidrule{2-8}
        & VOGP & 93.5 ± 32.6 & 0.93 ± .04 & 28.2 ± 3.8 & 0.96 ± .08 & 18.3 ± 2.6 & 0.99 ± .03 \\
        
        \midrule
        
        \multirow[t]{3}{*}{LAC}
        & NE & 250 & 0.95 ± .02 & 250 & 0.93 ± .03 & 250 & 0.93 ± .05 \\
        & PaVeBa & 290.0 ± 4.2 & 1.00 ± .00 & 266.4 ± 2.1 & 0.99 ± .01 & 264.1 ± 3.7 & 0.98 ± .02 \\
        \cmidrule{2-8}
        & VOGP & 69.9 ± 13.7 & 1.00 ± .00 & 27.4 ± 3.8 & 0.99 ± .01 & 37.9 ± 6.5 & 0.99 ± .01 \\
        
        \midrule
        
        \multirow[t]{3}{*}{VS}
        & NE & 500 & 0.93 ± .03 & 500 & 0.95 ± .04 & 500 & 0.90 ± .10 \\
        & PaVeBa & 515.2 ± 3.1 & 0.90 ± .03 & 503.3 ± 1.7 & 0.94 ± .03 & 502.4 ± 1.8 & 0.90 ± .05 \\
        \cmidrule{2-8}
        & VOGP & 406.2 ± 158.9 & 0.93 ± .02 & 34.8 ± 10.2 & 0.77 ± .06 & 23.6 ± 3.9 & 0.87 ± .11 \\
    
        \midrule
        
        \multirow[t]{3}{*}{SnAr}
        & NE & 2000 & 0.89 ± .08 & 2000 & 0.89 ± .12 & 2000 & 0.86 ± .30 \\
        & PaVeBa & 2016.9 ± 5.7 & 0.91 ± .03 & 2005.2 ± 2.6 & 0.87 ± .10 & 2001.8 ± 1.7 & 0.92 ± .11 \\
        \cmidrule{2-8}
        & VOGP & 102.5 ± 15.2 & 0.97 ± .02 & 41.4 ± 3.8 & 0.87 ± .12 & 36.4 ± 3.7 & 1.00 ± .00 \\
        \bottomrule
    \end{tabular}}
    \caption{
        $\epsilon$-F1 score comparison of VOGP with NE and PaVeBa on several datasets. S.C. refers to sample complexities of the algorithms.
    }
    \label{tab:vogp_vector_comp}
\end{table*}

\subsection{Ablation study for continuous domain adaptation}

\begin{figure}
    \centering
    \vspace{-0.5cm}    
    \begin{subfigure}[b]{0.48\textwidth}
        \centering
        \includegraphics[width=\textwidth]{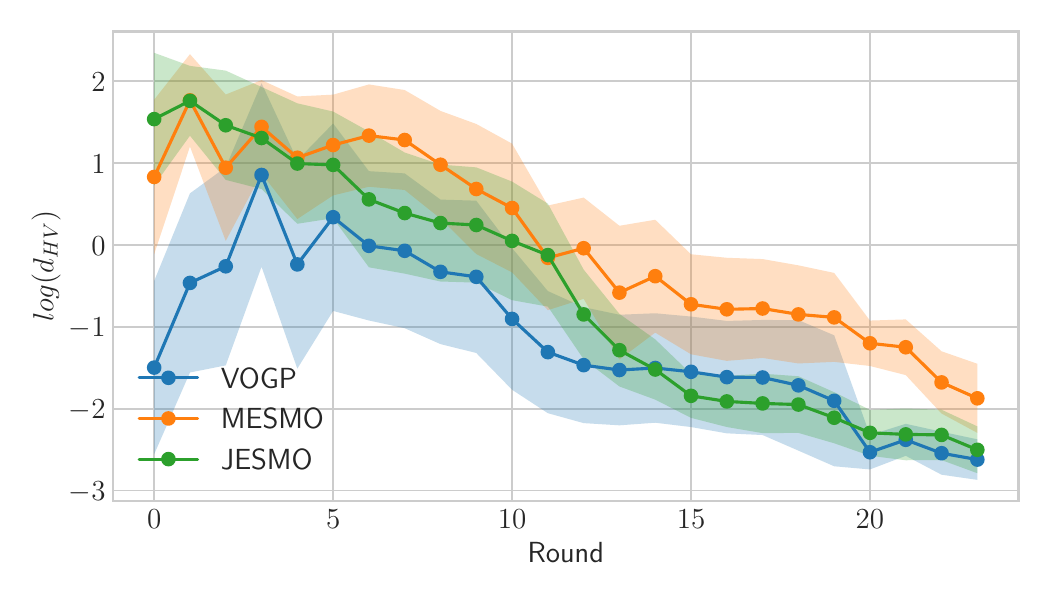}
        \label{fig:SR1}
    \end{subfigure}
    \begin{subfigure}[b]{0.48\textwidth}
        \centering
        \includegraphics[width=\textwidth]{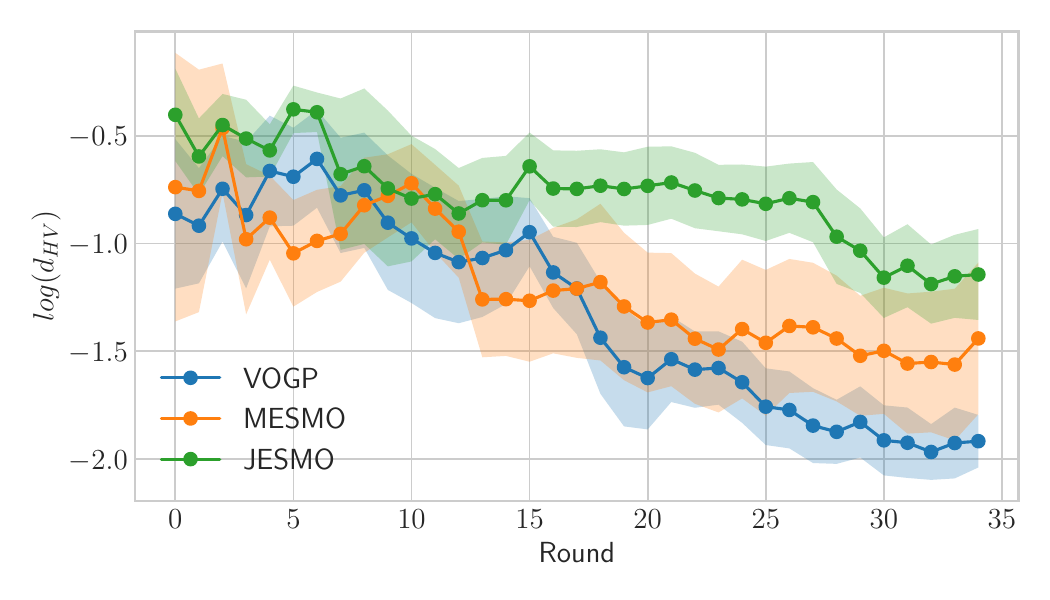}
        \label{fig:SR2}
    \end{subfigure}
    \vspace{-0.65cm}
    \caption{\label{fig:loghvd_vs_round}
    Mean logarithm hypervolume discrepancy vs. rounds on BCC (Left) and ZDT3 (Right) datasets, using right cone. Number of rounds is selected as the median S.C. Shaded regions correspond to half standard deviation.
    }
\end{figure}

Although theoretical bounds for VOGP are obtained for a finite set of designs $|\mathcal{X}|$, as an ablation, we implement a heuristic on top of VOGP to make it suitable for continuous problems. We adaptively discretize \citep{bubeck2011x, nika2021pareto} the continuous design space into cells, forming a tree structure with a hyperparameter that decides the maximum depth the tree can reach. We fix the maximum depth as $5$ in this case. We treat leaf nodes as designs while running VOGP. We split a leaf node into its $2^M$ child branches if it has high enough confidence and is not discarded (pruned) yet. We delay epsilon covering until the tree is maximally expanded and we run the remaining phases of VOGP as usual. Since the returned Pareto set after this process is not highly informative of what the Pareto front looks like in the continuous sense (as it is sparse), we treat VOGP like an exploration algorithm with adaptive maximum variance reduction as its acquisition function. Concretely, after the termination of VOGP we calculate the cone-dependent Pareto set using a uniform discretization over the posterior distribution of the GP. We do the same for MESMO and JES as well. We, then, calculate logarithm HV discrepancy for the found Pareto set for all algorithms. Since our theoretical confidence intervals assume finite domain, we consult to the literature for a confidence interval that is tight and has a theoretical underpinnings for continuous domains. We use the bound in \citet[Theorem~1]{fiedler2021practical}. Even though their reproducing kernel Hilbert space (RKHS) assumption does not coincide with our GP sample assumption, we found that this bound is empirically very sound. We run VOGP by taking the RKHS bound as $B=0.1$ and $\beta_t$ scaled by a factor $32$ for BCC and $48$ for ZDT3 problem. We present the results of this experiment in Table~\ref{tab:vogp_cont_comp} and Figure~\ref{fig:loghvd_vs_round}.

\begin{table*}[!t]
    \centering
    \resizebox{\columnwidth}{!}{
    \begin{tabular}{llcccccc}
        \toprule
        & & \multicolumn{2}{c}{Acute Cone} & \multicolumn{2}{c}{Right Cone} & \multicolumn{2}{c}{Obtuse Cone} \\
        \midrule
        Dataset & Alg. & S.C. & $\log\left(d_{\mathrm{HV}}\right)$ ($\downarrow$) & S.C. & $\log\left(d_{\mathrm{HV}}\right)$ ($\downarrow$) & S.C. & $\log\left(d_{\mathrm{HV}}\right)$ ($\downarrow$) \\
        \midrule
        \multirow[t]{3}{*}{BCC}
        & MESMO & & -0.96 ± 0.50 & & -1.52 ± 0.89 & & -1.37 ± 2.22 \\
        & JES & & -0.75 ± 0.73 & & -2.22 ± 0.61 & & -4.28 ± 1.35 \\
        \cmidrule{2-8}
        & VOGP & 168.6 ± 75.02 & 0.05 ± 0.72 & 27.5 ± 8.3 & -2.31 ± 0.72 & 23.3 ± 7.29 & -4.30 ± 1.30 \\

        \midrule

        \multirow[t]{3}{*}{ZDT3}
        & MESMO & & -2.35 ± 0.65 & & -1.59 ± 0.87 & & -1.12 ± 0.52 \\
        & JES & & -2.62 ± 0.71 & & -1.28 ± 0.89 & & -1.12 ± 0.97 \\
        \cmidrule{2-8}
        & VOGP & 152.0 ± 69.0 & -2.65 ± 0.67 & 64.8 ± 101.2 & -1.75 ± 0.67 & 38.9 ± 15.8 & -1.68 ± 0.96 \\
        \bottomrule
    \end{tabular}}
    \caption{
        Logarithm hypervolume discrepancy comparison of VOGP with MESMO and JES on BCC and ZDT3 problems in continuous domain. S.C. refers to sample complexities of the algorithms. Lower is better for both metrics.
    }
    \label{tab:vogp_cont_comp}
\end{table*}

\subsection{Adapting MO methods to vector optimization via linear transformations} \label{app:w_transform}

\begin{table}[ht]
\centering
\begin{tabular}{ccccc}
    \toprule
    N & Alg. & Time (sec.) & S.C. & $\epsilon$-F1 \\
    \midrule
    9 & VOGP & 22.38 & $28.50 \pm 3.8$ & $0.88\pm 0.09$ \\
    9 & MESMO & 104.26 & & $0.83 \pm 0.08$ \\
    9 & JES & DNF & & $-$ \\
    \addlinespace
    27 & VOGP & 59.68 & $28.30 \pm 3.20$ & $0.86 \pm 0.09$ \\
    27 & MESMO & 299.50 & & $0.82 \pm 0.06$ \\
    27 & JES & DNF & & $-$ \\
    \addlinespace
    81 & VOGP & 173.56 & $28.30 \pm 3.20$ & $0.86 \pm 0.09$ \\
    81 & MESMO & 879.76 & & $0.84 \pm 0.08$ \\
    81 & JES & DNF & & $-$ \\
    \bottomrule
\end{tabular}
\caption{
Comparison using different number of half spaces for ice cream cone approximation experiment. "DNF" means "Did not finish."
}
\label{tab:w_transform_ablation}
\end{table}

The cone ordering can be characterized by the usual componentwise ordering via the linear mapping induced by matrix $\boldsymbol{W}$. This can be confirmed by observing that 
\begin{equation}
    \boldsymbol{x} \preccurlyeq_C \boldsymbol{y} \iff \boldsymbol{W} (\boldsymbol{y}-\boldsymbol{x}) \geq \boldsymbol{0} \iff \boldsymbol{W} \boldsymbol{y} \geq \boldsymbol{W} \boldsymbol{x} \iff \boldsymbol{W} \boldsymbol{x} \leq \boldsymbol{W} \boldsymbol{y}~, \label{eq:coneright}
\end{equation}
where $\leq, \geq$ are componentwise orders (i.e. positive orthant order). A naive approach of adapting multi-objective optimization (MO) methods to vector optimization is using them per usual with the transformed objective function $\boldsymbol{W} \circ \boldsymbol{f}$. Please note that this linear transformation is $\mathbb{R}^M \rightarrow \mathbb{R}^N$, where $M$ is the number of objectives and $N$ is the number of halfspaces that define the polyhedral cone. In the paper, the concept of vector optimization is demonstrated mostly for the case $M=2$ since it is possible to visualize this case easily. Note that when $M=2$, we necessarily have $N=2$ for a solid pointed ordering cone $C$. However, when $M\geq 3$, in general, the number of halfspaces $N$ can be (much) larger than the number of objectives $M$. Hence, applying componentwise optimization methods (e.g. JES) to the transformed predicted objective values results in running these methods in the $N$-dimensional objective space (hence, an $N$-objective GP), which increases the computational cost of these methods dramatically, rendering them ineffective.

In this section we illustrate these challenges, particularly when the objective space dimension is modest (for instance, $M=3$) yet the complexity introduced by a large number of halfspaces defining the polyhedral cone ($N>M$) is significant—a scenario common in practice (e.g. P1 of \cite{hunt2010relative}). This is a crucial distinction because it underscores a unique advantage of VOGP: its ability to efficiently navigate the solution space in vector optimization problems, even when the the number of halfspaces defining the polyhedral cone (indicated by $N$) is high.

Consider the ice cream cone (expressed by the inequality $x_3 \geq \sqrt{x_1^2 + x_2^2}$) for $M=3$, and take its rotated version whose symmetry axis lies in the ray in $\mathbb{R}^3_+$ given by the equation $x_1=x_2=x_3$. Notice that the rotated ice cream cone is not polyhedral but one can construct a polyhedral approximation $C$ that is defined as the intersection of $N\geq 3$ halfspaces. The cone in Figure~\ref{fig:six} is a polyhedral approximation of the ice cream cone with $N=6$. We conduct an experiment on VS dataset using different $\boldsymbol{W}$ matrices that yield polyhedral approximations of the rotated ice cream cone with an increasing number ($N$) of halfspaces. We conduct the experiments on a discrete domain to limit acquisition evaluation of MO methods and since it conveys the message well enough. We use the original VOGP with $\beta_t$ set as the same as the continuous domain ablation. It can be seen that the computation times of MO methods (MESMO and JES) increase drastically with the number of halfspaces ($N$), highlighting that the formulation suggested is not really practical. On the contrary, we are able to obtain the results for $N=81$ with VOGP in reasonable computation time.

For $N=9$ (and also $N=27$ and $N=81$), while running JES algorithm, we observed hundreds of gigabytes of RAM usage and the algorithm raised out-of-memory error in our system with 128GB of RAM and 256GB of swap area.

\subsection{Ablation study for unknown GP hyperparameters} \label{app:unkown}

In some situations, neither the correct hyperparameters are inaccessible to us, nor do we possess reliable initial estimates for them. In such situations, even though the initial hyperparameters are not quite correct, a common approach is to gradually learn and adjust the hyperparameters as more queries get included in observations. So, in this ablation study we start with a random set of GP hyperparameters and after each evaluation phase, we update them by a maximum likelihood estimation(MLE) using the observations made so far.

Normally, after VOGP performs discarding and Pareto identification phases, it does not check the validity of decisions made in these phases in the subsequent rounds. This is because given that the GP model is accurate, these decisions are guaranteed to be good enough with high probability. However, in the unknown parameters setting, the parameters are yet to be meaningful in the earlier rounds. Hence, the decisions made in the earlier rounds can be erroneous. To adapt VOGP to this setting, at the end of each round, we reset the sets $\mathcal{S}_t$ and $\mathcal{P}_t$. With this adaptation, VOGP only terminates when all the designs are either discarded or moved to $\mathcal{P}_t$ in a single round. This ultimately allows VOGP to defer its decisions to when the kernel hyperparameters had been learned with the cost of doing repetitive work. We do not alter the rest of the algorithm.

\begin{table*}[!ht]
\centering
\begin{tabular}{lcc}
    \toprule
    Dataset & S.C. & $\epsilon$-F1 \\
    \midrule
    BC & 117.10 ± 20.27 & 0.99 ± .01 \\
    VS & 555.10 ± 147.16 & 1.00 ± .00 \\
    SnAr & 126.60 ± 19.77 & 0.96 ± .01 \\
    LAC & 99.70 ± 18.63 & 1.00 ± .00 \\
    \bottomrule
\end{tabular}
\caption{
    $\epsilon$-F1 scores of VOGP on several datasets under unknown hyperparameters setting. We only report results for acute cone. S.C. refers to sample complexity.
}
\label{tab:vogp_unknown_ablation}
\end{table*}

The results seem to improve over the known hyperparameters setting. We speculate this is because two things: (i) sample compexity of VOGP increases and (ii) these problems are not actually GP samples. First, sample complexity of VOGP is likely to be increased because it needs a high level of confidence of all designs to terminate in the span of a single round. Second, under the known hyperparameters setting, when MLE is done before the optimization process to find the "correct" hyperparameters, overall smoothness of the whole problem space is learned. Instead, if we could focus on and learn the local smoothness around the Pareto set, it would yield better results. Indeed, we think that this is what is happening when we start with unknown hyperparameters setting and learn them iteratively as VOGP focuses on identifying the Pareto set.

\subsection{Discussion of numerical results}
\rev{Table \ref{tab:vogp_vector_comp} demonstrates VOGP's high sample efficiency compared to the state-of-the-art methods in vector optimization. Specifically, we observe $\sim18.1\times$ and $\sim18.3\times$ lower sample complexity compared with NE and PaVeBa, respectively. These results are calculated as the mean of ratios of our method's average samples to those of NE's and PaVeBa's for each dataset/cone configuration in Table \ref{tab:vogp_vector_comp}. This drastic difference in sample complexity is possible since the theory of VOGP, specifically the use of surrogate GPs, allows for sequential decision making to happen after each observation, whereas other methods make decisions less frequently.} Despite its lower sample complexity, VOGP still outperforms the other methods in most cases. Table \ref{tab:vogp_cont_comp} and Figure \ref{fig:loghvd_vs_round} illustrate VOGP's robust performance when extended to continuous domains using adaptive discretization. VOGP remains competitive with the state-of-the-art information theoretic acquisition methods. Additionally, Table \ref{tab:vogp_unknown_ablation} shows that VOGP can be adapted to unknown hyperparameter settings and still perform very effectively. 
Most notably, Table \ref{tab:w_transform_ablation}, which uses linear transformations, emphasizes that VOGP or vector optimization in general is non-trivial. These results showcases the effectiveness of VOGP and its adaptability to different scenarios.

\section{Conclusion}
In this work we studied the problem of black-box vector optimization with Gaussian process bandits. We proposed VOGP, a sample efficient vector optimization algorithm, and provided success guarantees and upper bounds on its sample complexity. To the best of our knowledge, this is the first work that provides theoretical guarantees for black-box vector optimization with Gaussian process bandits. Through extensive experiments and ablations on various datasets and ordering cones, we showed our algorithms’ effectiveness over the existing works on vector optimization and straightforward adaptations of the MO methods. \rev{A limitation of our approach is that the theoretical guarantees are established under the assumption that the underlying surrogate model is a Gaussian process \revf{ and its hyperparameters are known a priori}. Extending the theoretical analysis \revf{to unknown hyperparameters or} to more general model classes, such as Bayesian neural networks remains an open challenge. Future research could explore such extensions, potentially broadening the applicability of vector optimization in settings where alternative surrogate models offer advantages in scalability or expressiveness. Other} interesting future research directions include providing theoretical guarantees for the problem in continuous spaces and developing counterparts of information theoretic acquisition functions for vector optimization.

\backmatter

\newpage
\begin{appendices}

\newpage
\section{Table of notation}\label{secA1}

Below, we present a table of symbols that are used in the proofs.

\begin{centering}
\begin{tabular}{ll}
\toprule
Symbol & Description \\
\midrule
$\mathcal{X}$ & The design space \\[0.2cm]
$M$ & The dimension of the objective space \\[0.2cm]
$\mathbb{S}^{M-1}$ & The unit sphere in $\mathbb{R}^M$ \\[0.2cm]
$C$ & The polyhedral ordering cone\\[0.2cm]
$\zeta$ & The ordering complexity of the cone $C$ \\[0.2cm]
$\bs{f}$ & The objective function \\[0.2cm]
$\bs{\epsilon}$ & Accuracy level given as input to the algorithm \\[0.2cm]
$\bs{y}_{[t]}$ & Vector that represents the first $t$ noisy observations where $\bs{y}_{[0]} = \emptyset$. \\[0.2cm]
$\bs{\mu}_t(x)$ & The posterior mean of design $x$ at round $t$ whose $j^\text{th}$ component is $\mu_t^j\left(x\right)$ \\[0.2cm]
$\bs{\sigma}_t(x)$ & The posterior variance of design $x$ at round $t$ whose $j^\text{th}$ component is $\sigma_t^j\left(x\right)$ \\[0.2cm]
$\beta_t$ & The confidence term at round $t$\\[0.2cm]
$\mathcal{P}_t$ & The predicted Pareto set of designs at round $t$\\[0.2cm]
$\mathcal{S}_t$ & The undecided sets of designs at round $t$\\[0.2cm]
$\hat{\mathcal{P}}$ & The estimated Pareto set of designs returned by VOGP \\[0.2cm]
$P^*$ & The set of true Pareto optimal designs \\ [0.2cm]
$P_\theta^*$ & The set of true Pareto optimal designs when $M=2$ and $C=C_{\theta}$\\ [0.2cm]
$\mathcal{A}_t$ & $\text { The union of sets } \mathcal{S}_t \text { and } \mathcal{P}_t \text { at the beginning of round } t$\\[0.2cm]
$\mathcal{W}_t$ & $\text { The union of sets } \mathcal{S}_t \text { and } \mathcal{P}_t \text { at the end of the discarding phase of round } t$ \\[0.2cm]
$\bs{Q}_t\left(x\right)$ & The confidence hyperrectangle associated with design $x$ at round $t$ \\[0.2cm]
$\bs{R}_t(x)$ & The cumulative confidence hyperrectangle associated with design $x$ at round $t$ \\[0.2cm]
$x_t$ & The design evaluated at round $t$ \\[0.2cm]
$\omega_t(x)$ & The diameter of the cumulative confidence hyperrectangle of design $x$ at round $t$ \\[0.2cm]
$\overline{\omega}_t$ &  The maximum value of $\omega_t(x)$ over all active designs $x$ at round $t$ \\[0.2cm]
$m(x, x^\prime)$ & $\inf \left\{s \geq 0 \mid \exists \bs{u} \in \mathbb{B}(1) \cap C: \bs{f}(x)+s \bs{u} \notin \bs{f}(x^\prime)-\operatorname{int}(C)\right\}$ \\[0.2cm]
$\gamma_t$ & The maximum information that can be gained about $\bs{f}$ in $t$ evaluations \\[0.2cm]
$t_s$ & The round in which VOGP terminates \\[0.2cm]
\hline
\end{tabular}
\end{centering}

\newpage

\section{Analysis for a Matérn kernel}\label{sec:mat}
The algorithm takes at most 
\begin{scriptsize}
\[
\mathcal{O}\left(\frac{ d_C^2(1)}{\epsilon^2}\left({{\frac{2\nu}{2\nu+D}}^{-\frac{4\nu+D}{2\nu+D}}}\right)  M^2  \sigma^2 \eta \cdot \ln^{\frac{4 \nu+D}{2 \nu+D}} \left({\left(\frac{\pi^2 M|\mathcal{X}|}{3 \delta}\right)}^{\left(\frac{\nu}{2\nu+D}\right)}\cdot \left({{\frac{2\nu}{2\nu+D}}^{-\frac{4\nu+D}{2\nu+D}}}\right) \frac{ d_C^2(1)}{\epsilon^2}  M^2  \sigma^2 \eta\right)
\right)
\]
\end{scriptsize}
samples, when it is run using a Matérn kernel where $\nu$ is the smoothness parameter.

\textit{Proof.}
By Lemma \ref{lem:terminationcondition_withdirection}, we have to show that taking 
\newline
\scalebox{0.75}{
\begin{minipage}{1.0\linewidth}
\begin{align*}
    t= \left(\frac{32 d_C^2(1)}{\epsilon^2}\left({{\frac{2\nu}{2\nu+D}}^{-\frac{4\nu+D}{2\nu+D}}}\right)  M^2  \sigma^2 \eta \phi \cdot \ln^{\frac{4 \nu+D}{2 \nu+D}} \left({\left(\frac{\pi^2 M|\mathcal{X}|}{3 \delta}\right)}^{\left(\frac{\nu}{2\nu+D}\right)}\cdot \left({{\frac{2\nu}{2\nu+D}}^{-\frac{4\nu+D}{2\nu+D}}}\right) \frac{32 d_C^2(1)}{\epsilon^2}  M^2  \sigma^2 \eta \phi\right)\right)^{\frac{2\nu+D}{2\nu}}
\end{align*}
\end{minipage}
}

satisfies $ \frac{8\beta_{t}  \sigma^2 \eta M \gamma_{t} }{t}  < \frac{\epsilon^2}{d_C^2(1)}$.
\vspace{0.2in}
\newline $\phi$ is the multiplicative constant that comes from $\mathcal{O}(\cdot)$ notation of the bound on single output GP's information gain. We will first upper-bound the LHS of this inequality and show that its smaller than or equal to the RHS of the inequality. We are using an Matérn kernel kernel $\tilde{k}$ for the design space. Then, by Lemma~\ref{lem:siribound} and the bounds on maximum information gain established in \cite{vakilibounds}, we have 

\[
      I(\boldsymbol{y}_{[t]}, \boldsymbol{f}_{[t]}) \leq  M \cdot \mathcal{O}\left(T^{\frac{D}{2 \nu+D}} \ln^ \frac{2 \nu}{2 \nu+D}(T)\right) 
    \implies \gamma_t \leq  M \cdot \mathcal{O}\left(T^{\frac{D}{2 \nu+D}} \ln^ \frac{2 \nu}{2 \nu+D}(T)\right)~.
\]
Notice that in Theorem \ref{thm:main}, as $\epsilon$ goes to 0, $T$ goes to infinity. Therefore, we can use the bounds on maximum information gain established in \cite{srinivas2012information}.  We have $\beta_t =\ln (\frac{M\pi^2 |\mathcal{X}| t^2}{3\delta })$.

\begin{align*}
    \frac{8 \beta_t \sigma^2 \eta M \gamma_t}{t} &= \frac{8 \ln (\frac{M\pi^2 |\mathcal{X}| t^2}{3\delta }) \sigma^2 \eta M \gamma_t}{t} \\
     &\leq \frac{8 \ln (\frac{M\pi^2 |\mathcal{X}| t^2}{3\delta }) \sigma^2 \eta M^2 \phi t^{\frac{D}{2 \nu+D}} \ln^ \frac{2 \nu}{2 \nu+D}(t)}  {t} \\
    & =  \frac{16 \ln \left(\sqrt{\frac{M|\mathcal{X}|}{3 \delta}} \cdot t\pi\right) \sigma^2 \eta \phi M^2 \ln^{\frac{2 \nu}{2 \nu+D}}(t)}{t^{\frac{2\nu}{2\nu+D}}} \\
    &<  \frac{16 \ln^{\frac{4 \nu+D}{2 \nu+D}} \left(\sqrt{\frac{M|\mathcal{X}|}{3 \delta}} \cdot t\pi\right) \sigma^2 \eta \phi M^2 }{t^{\frac{2\nu}{2\nu+D}}} \\
    & =  \frac{16 \alpha \ln^{\frac{4 \nu+D}{2 \nu+D}} \left(\sqrt{\frac{M|\mathcal{X}|}{3 \delta}} \cdot t\pi\right) \sigma^2 \eta \phi M^2 }{\alpha t^{\frac{2\nu}{2\nu+D}}} \\
    & =  \frac{16 {(\alpha^{\frac{2\nu+D}{4\nu+D}})}^{\frac{4 \nu+D}{2 \nu+D}}\ln^{\frac{4 \nu+D}{2 \nu+D}} \left(\sqrt{\frac{M|\mathcal{X}|}{3 \delta}} \cdot t\pi\right) \sigma^2 \eta \phi M^2 }{\alpha t^{\frac{2\nu}{2\nu+D}}} \\
    &=  \frac{16  \ln^{\frac{4 \nu+D}{2 \nu+D}} \left({\frac{\pi^2 M|\mathcal{X}|}{3 \delta}}^{(\frac{\alpha^{\frac{2\nu+D}{4\nu+D}}}{2})} \cdot t^{(\alpha^{\frac{2\nu+D}{4\nu+D}})}\right) \sigma^2 \eta \phi M^2 }{\alpha t^{\frac{2\nu}{2\nu+D}}} \\
\end{align*}
The strict inequality follows from $M\geq 1, \delta \in (0,1), \abs{\mathcal{X}} \geq 1$ and $\pi > \sqrt{3}$. Now, we can plug in $\alpha = {{\frac{2\nu}{2\nu+D}}^{\frac{4\nu+D}{2\nu+D}}}$ and after cancellations we get:

\begin{align*}
    &=  \frac{16\left({{\frac{2\nu}{2\nu+D}}^{-\frac{4\nu+D}{2\nu+D}}}\right)  \ln^{\frac{4 \nu+D}{2 \nu+D}} \left({\left(\frac{\pi^2 M|\mathcal{X}|}{3 \delta}\right)}^{\left(\frac{\nu}{2\nu+D}\right)} \cdot t^{\left(\frac{2\nu}{2\nu+D}\right)}\right) \sigma^2 \eta \phi M^2 }{ t^{\frac{2\nu}{2\nu+D}}} \\
\end{align*}

Now, we can plug in

\scalebox{0.8}{
\begin{minipage}{1.0\linewidth}
\begin{align*}
    t^{\frac{2\nu}{2\nu+D}}= \frac{32 d_C^2(1)}{\epsilon^2}\left({{\frac{2\nu}{2\nu+D}}^{-\frac{4\nu+D}{2\nu+D}}}\right)  M^2  \sigma^2 \eta \phi \cdot \ln^{\frac{4 \nu+D}{2 \nu+D}} \left({\left(\frac{\pi^2 M|\mathcal{X}|}{3 \delta}\right)}^{\left(\frac{\nu}{2\nu+D}\right)}\cdot \left({{\frac{2\nu}{2\nu+D}}^{-\frac{4\nu+D}{2\nu+D}}}\right) \frac{32 d_C^2(1)}{\epsilon^2}  M^2  \sigma^2 \eta \phi\right)
\end{align*}
\end{minipage}
}

to the last expression:

\scalebox{0.6}{
\begin{minipage}{1.0\linewidth}
    \begin{align*}
    &=  \frac{16\left({{\frac{2\nu}{2\nu+D}}^{-\frac{4\nu+D}{2\nu+D}}}\right)  \ln^{\frac{4 \nu+D}{2 \nu+D}} \left({\left(\frac{\pi^2 M|\mathcal{X}|}{3 \delta}\right)}^{\left(\frac{\nu}{2\nu+D}\right)}\cdot \frac{32 d_C^2(1)}{\epsilon^2}\left({{\frac{2\nu}{2\nu+D}}^{-\frac{4\nu+D}{2\nu+D}}}\right)  M^2  \sigma^2 \eta \phi \cdot \ln^{\frac{4 \nu+D}{2 \nu+D}} \left({\left(\frac{\pi^2 M|\mathcal{X}|}{3 \delta}\right)}^{\left(\frac{\nu}{2\nu+D}\right)}\cdot \left({{\frac{2\nu}{2\nu+D}}^{-\frac{4\nu+D}{2\nu+D}}}\right) \frac{32 d_C^2(1)}{\epsilon^2}  M^2  \sigma^2 \eta \phi\right)\right) \sigma^2 \eta \phi M^2 }{ \frac{32 d_C^2(1)}{\epsilon^2}\left({{\frac{2\nu}{2\nu+D}}^{-\frac{4\nu+D}{2\nu+D}}}\right)  M^2  \sigma^2 \eta \phi \cdot \ln^{\frac{4 \nu+D}{2 \nu+D}} \left({\left(\frac{\pi^2 M|\mathcal{X}|}{3 \delta}\right)}^{\left(\frac{\nu}{2\nu+D}\right)}\cdot \left({{\frac{2\nu}{2\nu+D}}^{-\frac{4\nu+D}{2\nu+D}}}\right) \frac{32 d_C^2(1)}{\epsilon^2}  M^2  \sigma^2 \eta \phi\right)} \\
    & = \frac{\ln^{\frac{4 \nu+D}{2 \nu+D}} \left({\left(\frac{\pi^2 M|\mathcal{X}|}{3 \delta}\right)}^{\left(\frac{\nu}{2\nu+D}\right)}\cdot \frac{32 d_C^2(1)}{\epsilon^2}\left({{\frac{2\nu}{2\nu+D}}^{-\frac{4\nu+D}{2\nu+D}}}\right)  M^2  \sigma^2 \eta \phi \cdot \ln^{\frac{4 \nu+D}{2 \nu+D}} \left({\left(\frac{\pi^2 M|\mathcal{X}|}{3 \delta}\right)}^{\left(\frac{\nu}{2\nu+D}\right)}\cdot \left({{\frac{2\nu}{2\nu+D}}^{-\frac{4\nu+D}{2\nu+D}}}\right) \frac{32 d_C^2(1)}{\epsilon^2}  M^2  \sigma^2 \eta \phi\right)\right) }{ 2\ln^{\frac{4 \nu+D}{2 \nu+D}} \left({\left(\frac{\pi^2 M|\mathcal{X}|}{3 \delta}\right)}^{\left(\frac{\nu}{2\nu+D}\right)}\cdot \left({{\frac{2\nu}{2\nu+D}}^{-\frac{4\nu+D}{2\nu+D}}}\right) \frac{32 d_C^2(1)}{\epsilon^2}  M^2  \sigma^2 \eta \phi\right)} \cdot \frac{\epsilon^2}{  d_C^2(1)}\\
\end{align*}
\end{minipage}
}

To show that the last expression is smaller than or equal to $\frac{\epsilon^2}{  d_C^2(1)}$, we need to show that 
\[
\ln^{\frac{4 \nu+D}{2 \nu+D}}\left(B \ln^{\frac{4 \nu+D}{2 \nu+D}}(B)\right) \leq 2 \ln^{\frac{4 \nu+D}{2 \nu+D}}(B)=\ln^{\frac{4 \nu+D}{2 \nu+D}}(B ^{\left( 2^{\frac{2\nu+D}{4\nu+D}} \right)})
\]
where $B =\left({\left(\frac{\pi^2 M|\mathcal{X}|}{3 \delta}\right)}^{\left(\frac{\nu}{2\nu+D}\right)}\cdot \left({{\frac{2\nu}{2\nu+D}}^{-\frac{4\nu+D}{2\nu+D}}}\right) \frac{32 d_C^2(1)}{\epsilon^2}  M^2  \sigma^2 \eta \phi\right)$ for convenience. This holds because $\ln^{\frac{4 \nu+D}{2 \nu+D}} B \leq B ^{\left( 2^{\frac{2\nu+D}{4\nu+D}}-1 \right)}$ for sufficiently large  $B$. Hence, the proof is complete.

\newpage

\section{Convergence plots for Table \ref{tab:vogp_vector_comp}} \label{supp:exp}

\begin{figure}[h]
    \centering  
    \includegraphics[width=\textwidth]{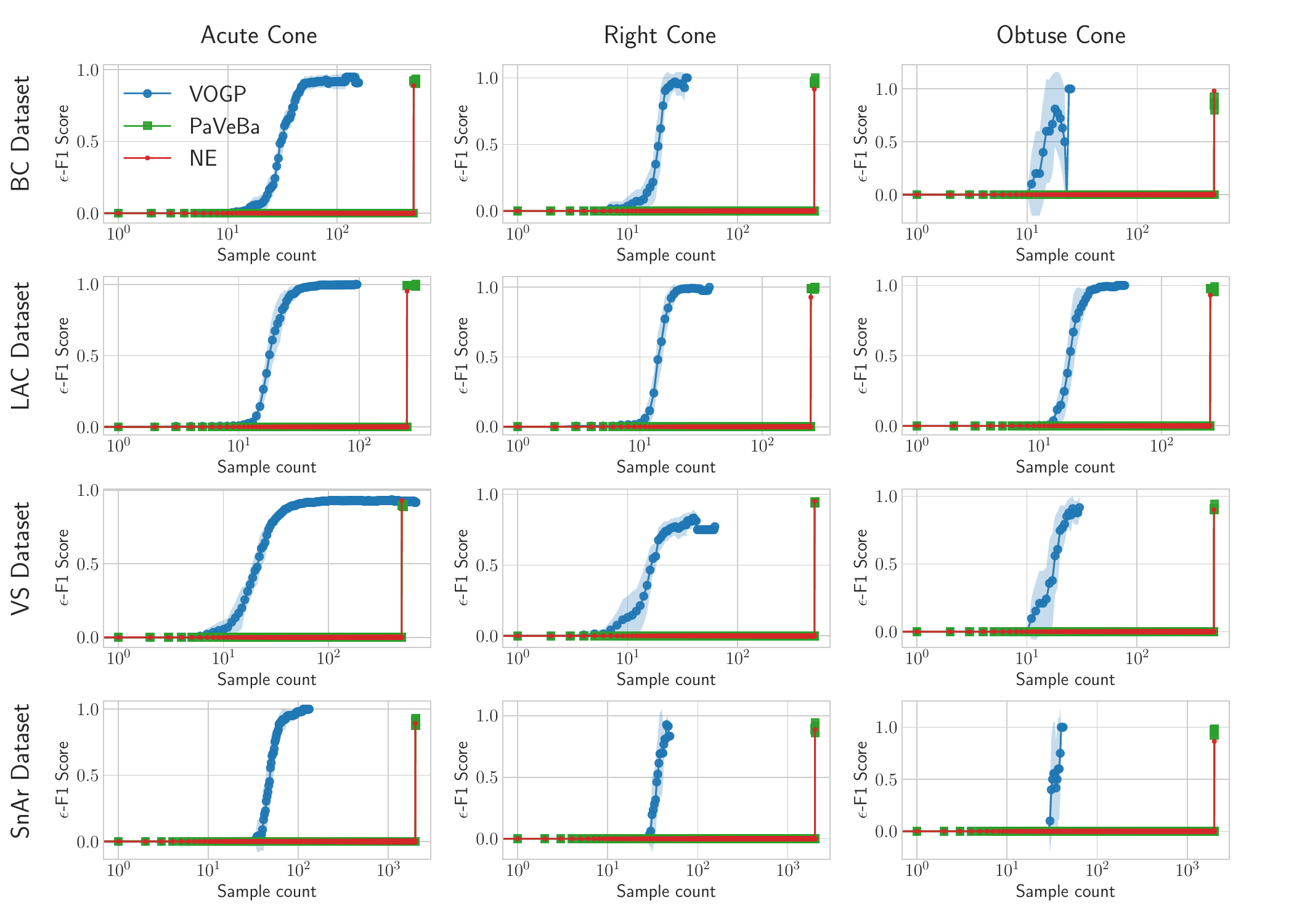}
    \caption{\label{fig:vogp_vector_comp_conv}
    \rev{
    Convergence plots of vector optimization methods, presenting $\epsilon$-F1 scores over sample counts (logarithmic scale is used for sample counts).
    }
    }
\end{figure}

Figure \ref{fig:vogp_vector_comp_conv} provides convergence plots of vector optimization methods, presenting $\epsilon$-F1 scores over sample counts. The experiment setup is exactly same as Table \ref{tab:vogp_vector_comp}. Each subfigure corresponds to a specific dataset and a specific ordering cone, indicated in the subfigure titles. \rev{In these plots, we present average $\epsilon$-F1 scores over sample counts of different iterations. Naturally, for each sample count, only the iterations that reach that specific sample count is considered. Therefore, note that some plots illustrate only some iterations around the end.} The Naive Elimination and PaVeBa algorithms require a large number of samples before making decisions, which leads to stagnant regions in the plots. In contrast, VOGP makes decisions at each round, resulting in a smoother decision trajectory and superior sample efficiency.

\section{Acknowledgements}
This work was supported by the Scientific and Technological Research Council of Türkiye (TÜBİTAK) Grant No 121E159. İ.O.K. and Y.C.Y were also supported by Türk Telekom as part of 5G and Beyond Joint Graduate Support Programme coordinated by Information and Communication Technologies Authority. C.T. acknowledges support by the Turkish Academy of Sciences Distinguished Young Scientist Award Program (TÜBA-GEBİP-2023) and the Scientific and Technological Research Council of Türkiye (TÜBİTAK) 2024 Incentive Award.
\section{Declarations} 

\textbf{Competing Interests} The authors declare that there are no competing interests.

\noindent
\textbf{Code Availability}
The code is available at \url{https://github.com/Bilkent-CYBORG/VOGP}.

\noindent
\textbf{Data Availability}
All data supporting the findings of this study is available at \url{https://github.com/Bilkent-CYBORG/VOGP}.

\noindent
\textbf{Author Contributions}
İ.O.K. authored most of the manuscript and derived the theoretical contributions. He also took part in development of the VOGP algorithms. İ.O.K. contributed in designing and executing the experiments and figure preparation. Y.C.Y. authored the experimental section, conducted all experiments, and significantly contributed to the manuscript editing and figure preparation. Ç.A. and C.T. provided critical oversight, verified theoretical proofs, and guided the strategic direction of the project. Both also and contributed to the manuscript's writing and editing, and conceptualized the idea.

\noindent
\textbf{Ethics Approval} Not applicable.

\noindent
\textbf{Consent to Participate} Not applicable.

\end{appendices}

\bibliography{ref_vogp}

\end{document}